\documentclass{article}

\PassOptionsToPackage{numbers}{natbib}

\usepackage[main, final]{neurips_2026}

\usepackage[utf8]{inputenc} 
\usepackage[T1]{fontenc}    
\usepackage{hyperref}       
\usepackage{url}            
\usepackage{booktabs}       
\usepackage{amsfonts}       
\usepackage{nicefrac}       
\usepackage{microtype}      
\usepackage{xcolor}         
\usepackage{amsmath}
\usepackage{amssymb}
\usepackage{graphicx}
\usepackage[table]{xcolor}
\usepackage{wrapfig}
\usepackage{multirow}
\usepackage{mdframed}

\newtheorem{definition}{Definition}
\usepackage{caption}
\usepackage{fontawesome5}

\title{The Commit-Abstain Circuit: Why Language Models Hallucinate Instead of Abstaining}

\author{%
  \textbf{Vy Nguyen\textsuperscript{1}} \quad
  \textbf{Ziqi Xu\textsuperscript{1}} \quad
  \textbf{Jeffrey Chan\textsuperscript{1}} \quad
  \textbf{Estrid He\textsuperscript{1}} \\
  \textbf{Feng Xia\textsuperscript{1}} \quad
  \textbf{Renqiang Luo\textsuperscript{2}} \quad
  \textbf{Erik Cambria\textsuperscript{3}} \quad
  \textbf{Xiuzhen Zhang\textsuperscript{1,*}} \\[4pt]
  \textsuperscript{1}RMIT University \quad
  \textsuperscript{2}Jilin University \quad
  \textsuperscript{3}Nanyang Technological University \\[5pt]
  {\ttfamily s3964786@student.rmit.edu.au, cambria@ntu.edu.sg, lrenqiang@jlu.edu.cn} \\
  {\ttfamily \{ziqi.xu, jeffrey.chan, estrid.he, feng.xia, xiuzhen.zhang\}@rmit.edu.au}
    \vspace{2mm}
  \\
  \textsuperscript{*}Corresponding author \\[6pt]
  \href{https://github.com/vnht/commit-abstain-circuit}
  {\faGithub\ \texttt{github.com/vnht/commit-abstain-circuit}}
}

\begin{document}
\maketitle

\begin{abstract}
Language models (LMs) often hallucinate by committing to confident answers rather than abstaining, even when they do not have enough information to answer reliably. A large body of existing work mitigates hallucination through detection or abstention mechanisms, but leaves open how models internally arrive at the decision to commit or abstain in the first place. We study this decision through mechanistic analysis, framing hallucination as unsupported commitment: the model commits despite exhibiting signals of unanswerability. Using causal gating, we identify a \textbf{C}ommit-\textbf{A}bstain \textbf{C}ircuit (CAC), a sparse, causally localised subset of attention heads and MLP sublayers underlying this decision. Across ten LMs (3B--14B) from five families and three benchmarks, the CAC exhibits a recurring \textit{accumulate-yet-undercorrect} pattern: commitment-promoting components build up commitment in earlier layers, while abstention-promoting components act later as corrective signals that are often insufficient to overturn the accumulated commitment. Building on this finding, a lightweight policy trained on CAC activations improves decision accuracy by $12.2$ points over the model's intrinsic commit-abstain margin, reduces false abstentions by $2.5$ times, transfers to unseen benchmarks, and extends to larger models (27B--35B). The CAC is both diagnostic, clarifying how models overcommit, and practical, enabling improved abstention decisions.\end{abstract}

\section{Introduction}

Reliable language models (LMs) need two capabilities: answering when they can, and abstaining when they should, for example by responding ``I don't know'' or indicating that the available information is insufficient~\citep{must-be-taught-2024-Kapoor, wen-etal-2025-know}. Abstention, however, remains an open challenge: large-scale evaluations show that even frontier models fail to abstain reliably across forms of \textit{unanswerability} such as unknown futures, false premises, missing context, and underspecification, and that neither scaling nor extended reasoning reliably fixes this~\citep{kirichenko2025abstentionbench,muhamed-etal-2026-refusalbench,peng-etal-2025-unanswerability}. Training and evaluation regimes tend to reward guessing, pushing models toward hallucination rather than honest abstention~\citep{Kalai2026}.

A large body of work treats hallucination as a persistent limitation of LMs~\citep{LLM_Check,estimating-hallucination-rate,bang-etal-2025-hallulens}, emphasising detection and mitigation of hallucinations in model outputs~\citep{Farquhar2024, du2024haloscope, zhou2025hademif, ZhaoZ0R0LF025, LiXRLZZRX26,RenZLLLCCXXL26}, or guiding and training models to abstain reliably~\citep{feng-etal-2024-dont, zhang-etal-2024-r,2025-causal-abstention}. These approaches mitigate hallucination at the behavioural level but leave unexplained how the model internally arrives at hallucination.

Emerging work suggests that unanswerability information is already present in the model's internal states: linear probes decode it from hidden activations, and intervening on their directions changes abstention behaviour~\citep{lavi-etal-2026-detecting,simhi-etal-2025-trust,yu-etal-2024-mechanistic}. Mechanistic analyses identify entity-recognition circuits whose misfires produce hallucinations~\citep{ferrando2025know,lindsey2025biology}. However, neither line of work explains the model's failure to act: if unanswerability signals are already present, why does the model still answer confidently? Existing work studies representations and directions associated with hallucination, but does not localise which model components underlie the commit-abstain decision or account for how the model answers in the first place. We address this gap by asking how the model internally fails to produce abstention, framing hallucination as a control failure rather than a knowledge failure.

To this end, we define \emph{unsupported commitment} as the model's failure to abstain despite exhibiting signals of unanswerability, producing a substantive answer rather than deferring with expressions such as ``I don't know'' or ``The context is insufficient''. We ask two questions: (RQ1) Does a sparse subset of model components causally contribute to the model's commit-abstain decision, and (RQ2) Can an understanding of this subset support a better abstention policy than the model's default behaviour?

\begin{figure}[t]
    \centering
    \includegraphics[width=1\textwidth]{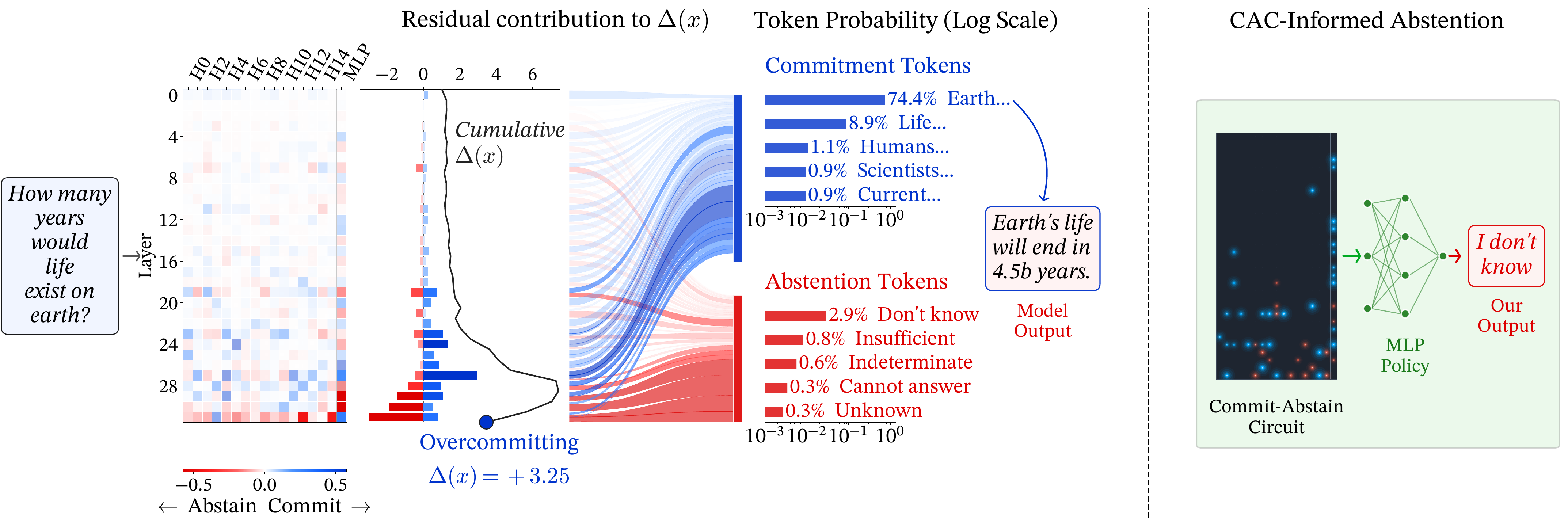}
    \caption{Unsupported commitment in Qwen 3.5 4B on an unanswerable KUQ question~\citep{amayuelas-etal-2024-knowledge}. \textbf{Left:} Per-component contributions to the commit-abstain margin $\Delta(x)$ (Definition~\ref{Commit-Abstain Margin}). Late-layer abstention signals fail to overturn earlier commitment ($\Delta(x)=+3.25$), leading to commitment. \textbf{Right:} A lightweight CAC-Informed MLP correctly triggers abstention.}
    \label{fig:main-example}
\end{figure}

We analyse the commit-abstain decision from the model's logits before any token is emitted, capturing its preference before autoregressive generation introduces path dependence~\citep{afzal-etal-2025-knowing,the-first-to-know-2025}. We quantify this preference as a \emph{commit-abstain margin} and decompose it over attention heads and multilayer perceptron (MLP) sublayers. We adapt causal head gating~\citep{nam2025causal} to the commit-abstain setting and identify a sparse, causally localised subset of the model's components that underlie this decision, termed the \textbf{C}ommit-\textbf{A}bstain \textbf{C}ircuit (CAC). Across ten LMs and three benchmarks, a recurring pattern emerges in the CAC: commitment-promoting components act in earlier layers and accumulate commitment, while abstention-promoting components act later as corrective signals that are often insufficient to overturn it. Figure~\ref{fig:main-example} illustrates this pattern on an unanswerable input: the abstention signal arises too late to overturn the accumulated commitment, and the model commits. This pattern provides a component-level account of unsupported commitment and motivates a lightweight policy that improves abstention beyond the default behaviour of the model. Our contributions are as follows:

\begin{itemize}
    \item \textbf{Problem Formulation.} We frame hallucination as \emph{unsupported commitment}, a failure of control rather than of knowledge, and introduce a commit-abstain margin to quantify the model's commit-abstain preference.
    
    \item \textbf{Mechanistic Finding.} We identify the \textbf{C}ommit-\textbf{A}bstain \textbf{C}ircuit (CAC), a sparse, causally localised subset of the model's attention heads and MLP sublayers (median 5.2\% of components) that underlie the commit-abstain margin, and reveal a recurring \textit{accumulate-yet-undercorrect} pattern, providing a component-level account of unsupported commitment.

    \item \textbf{Policy Improvement.} We train a lightweight policy on CAC activations that improves decision accuracy by $12.2$ points and reduces false abstentions by $2.5\times$ across ten models (3B--14B) from five families and three benchmarks, with gains transferring to unseen benchmarks and reproduced on larger models (27B--35B).
\end{itemize}

\section{Related Work}
\paragraph{Hallucination Detection and Abstention}
Hallucination, the generation of confident but unsupported content, is a core reliability failure in LMs~\citep{Huang2025HallucinationSurvey,Alansari2026Hallucination}. Two broad lines of work address this: \textit{detecting} hallucinations in outputs~\citep{Farquhar2024,manakul-etal-2023-selfcheckgpt} and \textit{abstaining} when answering would likely hallucinate~\citep{wen-etal-2025-know}. \textit{Detection} methods span three main families: output-level uncertainty, confidence, or entropy estimation~\citep{Farquhar2024,manakul-etal-2023-selfcheckgpt,ciosek2025hallucination,Chen2023HallucinationDetection}; evidence-based verification~\citep{li-etal-2024-dawn,hu-etal-2024-knowledge,Liu2025hallucination,niu-etal-2024-ragtruth}; and model-internal probing of hidden states, attention, or representations~\citep{zhang-etal-2025-icr,chen2024inside,du2024haloscope,niu2025robust}. \textit{Abstention} methods split into training-free approaches such as selective answering~\citep{lee2024selective,xin-etal-2021-art,Li2024ThinkTwice}, uncertainty control~\citep{kim-etal-2025-speak,vazhentsev-etal-2025-unconditional,nikitin2024kernel,yadkori-conformal-abstention,abbasi-yadkori2024to}, consensus~\citep{feng-etal-2024-dont,Liang2026agenthalluc,Wen2025Marvel}, and causal analysis~\citep{sun-etal-2025-causalabstain,2025-causal-abstention,Li2026CausalHallu, equivariance}; and training-based approaches using fine-tuning~\citep{cohen2024i,zhang-etal-2024-r,huang-etal-2025-alleviating,Shi2026Fine-Tuned,stengel-eskin2024lacie} or reinforcement learning~\citep{Wei2025TruthRL,Zhao2025AutomaticCurriculum,Xu2024Rejection}. Both lines treat hallucination as an established phenomenon to detect or mitigate, without explaining how the model internally arrives at hallucination. In contrast, we take a mechanism-first approach: we localise internal components that contribute to the commit-abstain decision, and use this understanding to inform improved abstention.

\paragraph{Mechanistic Interpretability}
Mechanistic work has identified sparse circuits underlying diverse model behaviours: induction heads copy from context~\citep{olsson2022context}, successor heads increment ordinals~\citep{gould2024successor}, distinct heads mediate contextual versus parametric knowledge~\citep{atlas-of-in-context,wu2025retrieval,sun2025redeep,ChengXLLYLL26}, and factual recall routes through MLP pathways~\citep{meng2022locating,geva-etal-2023-dissecting}. Closest to our setting is work on safety refusal, which is mediated by a sparse residual-stream sub-circuit~\citep{refusal-single-direction,Prakash2026BeyondImSorry}. Our focus is different: epistemic abstention asks whether the model knows enough to answer, not whether it should refuse a harmful request. We adapt causal gating~\citep{nam2025causal} to this setting, extending it to gate MLP sublayers alongside attention heads under a margin-based objective for the commit-abstain decision.
\paragraph{The Representation-Behaviour Gap}
Emerging work shows that models carry information that their generations fail to express. Probes and geometric analyses reveal signals of correctness, truthfulness, and unanswerability in hidden states~\citep{azaria-mitchell-2023-internal,burns2023discovering,liu2023cognitive,orgad2025llms,simhi-etal-2025-trust}, including signals present prior to generation~\citep{niu2025robust,the-first-to-know-2025,ni-etal-2025-towards,afzal-etal-2025-knowing}. Recent work further investigates how such signals fail to translate into behaviour, including entity-recognition mechanisms whose misfiring produces hallucination~\citep{ferrando2025know,lindsey2025biology}. These findings suggest a representation-behaviour gap: information relevant to whether the model should answer is present internally, yet the model may still produce an answer. Prior work detects these signals but has not characterised the internal components that underlie the commit-abstain decision. In contrast, we identify a sparse subset of attention heads and MLP sublayers and show that their component-level contributions retain unanswerability information that is partially lost when aggregated into the final commit-abstain margin.

\section{The Commit-Abstain Circuit}
\label{sec:formulation}

To characterise unsupported commitment, we need to measure the model's commit-abstain preference. We quantify this from the model's logits before autoregressive generation introduces path dependence. This design is supported by evidence that decision-relevant signals such as correctness and truthfulness are encoded prior to generation~\citep{orgad2025llms, afzal-etal-2025-knowing, the-first-to-know-2025}, and that such logits provide a reliable readout of model preferences~\citep{first-tokens-are-different, huang2024large, niu2025robust}.

We partition the vocabulary into an abstention set $\mathcal{A}$ and a commitment set $\mathcal{C}$. Tokens in $\mathcal{A}$ initiate deferral expressions, while $\mathcal{C}$ contains the remaining vocabulary. Table~\ref{tab:expression-families} provides examples of expressions initiated by each set. We describe the empirical construction of $\mathcal{A}$ in Section~\ref{sec:exp-settings}. Intuitively, $\Delta(x)$ compares the model's strongest immediate tendency to answer with its strongest tendency to abstain.

For input $x$, let $r_L(x) \in \mathbb{R}^d$ denote the residual state at the final input position. Following prior practice in direct logit attribution~\citep{meng2022locating, marks2025sparse}, we absorb the final normalisation into the unembedding, so the per-token logit is $z_t(x) = u_t^\top r_L(x) + b_t$, where $u_t \in \mathbb{R}^d$ is the unembedding vector for token $t$ and $b_t$ is its bias (zero if absent). We then define the commit-abstain margin:

\begin{definition}[Commit-Abstain Margin]\label{Commit-Abstain Margin}
Let $c^*(x) = \arg\max_{t \in \mathcal{C}}\, z_t(x)$ and $a^*(x) = \arg\max_{t \in \mathcal{A}}\, z_t(x)$ denote the highest-logit tokens in $\mathcal{C}$ and $\mathcal{A}$. The commit-abstain margin is
\begin{equation}
\Delta(x) = z_{c^*(x)}(x) - z_{a^*(x)}(x).
\end{equation}
$\Delta(x) \geq 0$ indicates a preference to commit; $\Delta(x) < 0$ indicates a preference to abstain.
\end{definition}

\paragraph{Remark 1}
\label{rmk:unanswerability-info}
Let $Y \in \{0, 1\}$ be the gold answerability label. If $\mathrm{AUROC}(\Delta(X), Y) > \tfrac{1}{2}$, the conditional distributions $P(\Delta(X) \mid Y=1)$ and $P(\Delta(X) \mid Y=0)$ differ, so $I(\Delta(X); Y) > 0$, where $I(\cdot;\cdot)$ denotes mutual information. By the data-processing inequality applied to the deterministic map $r_L(X) \mapsto \Delta(X)$, $I(r_L(X); Y) \geq I(\Delta(X); Y) > 0$. The residual stream therefore carries information about unanswerability whenever $\Delta(X)$ separates classes above chance, and thus $\Delta(X)$ provides a tractable summary of this internal information.

To study this internal mechanism, we exploit the additive structure of the residual stream. The final residual state $r_L(x)$ can be written as a sum of contributions from all attention heads and MLP sublayers. Under this formulation, $\Delta(x)$ decomposes into signed per-component contributions: positive contributions push the margin toward commitment, while negative contributions push it toward abstention. Let $d(x) = u_{c^*(x)} - u_{a^*(x)}$ denote the direction along which the residual stream contributes to $\Delta(x)$. Then,
\begin{equation}
\Delta(x) = \kappa(x) + \sum_{k=1}^{K} v_k(x),
\end{equation}
where $\kappa(x)$ collects bias and embedding terms, $k$ indexes attention heads and MLP sublayers, $c_k(x) \in \mathbb{R}^d$ is the output activation of component $k$, and $v_k(x) = d(x)^\top c_k(x)$ is its signed contribution projected onto $d(x)$. Per-component contributions are computed under the standard linearisation that the final normalisation factor is fixed to its value at the ungated forward pass~\citep{meng2022locating, marks2025sparse}.

The additive decomposition of $\Delta(x)$ suggests that the commit-abstain decision may be mediated by a subset of components, formalising RQ1. We define this subset as follows:

\begin{definition}[Commit-Abstain Circuit]
The \textbf{C}ommit-\textbf{A}bstain \textbf{C}ircuit (CAC) is a causally localised subset of attention heads and MLP sublayers that mediate the commit-abstain preference encoded by $\Delta(x)$.
\end{definition}

We localise and characterise the CAC empirically in Section~\ref{sec:localise-cac} and use it to design an improved abstention policy in Section~\ref{sec:policy-cac}.

\begin{table}[t]
\caption{Token sets used to define the commit-abstain margin $\Delta(x)$.}
\label{tab:expression-families}
\centering
\footnotesize
\renewcommand{\arraystretch}{0.7}
\begin{tabular}{p{0.47\textwidth}p{0.47\textwidth}}
\toprule
\textbf{Abstention set $\mathcal{A}$} & \textbf{Commitment set $\mathcal{C}$} \\
\midrule
``Don't know'', ``Insufficient evidence'', ``Unsure'', ``Unanswerable'', ``Ambiguous'', ``Beyond my knowledge'' &
All remaining, e.g.\ those initiating: ``According to...'', ``The answer is...'',
``Exactly'', ``Yes'', ``No'', ``In fact'' \\
\bottomrule
\end{tabular}
\end{table}

\section{Localisation of the Commit-Abstain Circuit}
\label{sec:localise-cac}

We address RQ1 by verifying that $\Delta(x)$ reflects signals of unanswerability, identifying the CAC as its component-level basis, and characterising how the CAC shapes the commit-abstain margin.
\subsection{Setup for CAC Localisation}
\label{sec:exp-settings}

\paragraph{Datasets}
We use three benchmarks spanning parametric recall, in-context retrieval, and complex reasoning: \textbf{KUQ}~\citep{amayuelas-etal-2024-knowledge}, which contains open-domain questions with no definitive answer; \textbf{SQuAD}~2.0~\citep{rajpurkar-etal-2018-know}, a reading comprehension dataset with adversarially constructed unanswerable contexts; and \textbf{MuSiQue}~\citep{trivedi-etal-2022-musique}, which includes multi-hop questions with a supporting document removed. A detailed description of the datasets is provided in App.~\ref{app:datasets}. We sample $1{,}000$ instances from each dataset. All supervised experiments use a $50/50$ train-test split.

\paragraph{Models}
We evaluate ten instruction-tuned LMs spanning five model families and two parameter scales: Qwen-3.5 (4B, 9B)~\citep{qwen3}, Gemma-3 (4B, 12B)~\citep{gemma3}, Llama-3.2 (3B) and Llama-3.1 (8B)~\citep{llama3}, Ministral-3 (3B, 14B)~\citep{ministral3}, and Phi-4-mini (4B) and Phi-4 (14B)~\citep{phi4}. All models use greedy decoding with $\texttt{max\_new\_tokens}=256$ and are run on a NVIDIA H100 GPU.

\paragraph{Construction of $\mathcal{A}$}
For each model, zero-shot inference is performed on a separate pool of $1{,}000$ unanswerable instances disjoint from the train and test splits, and the initial token of each response that matches a canonical deferral pattern is recorded. These tokens are pooled across models, deduplicated, disambiguated, and manually validated to ensure that they initiate epistemic deferral. We provide the full procedure, token lists, and a robustness analysis to the choice of $\mathcal{A}$ in App.~\ref{app:token-families}.

\subsection{Unanswerability Signals in the Commit-Abstain Margin}
\label{sec:unanswerability-signals}

\begin{figure*}[t]
\centering
\begin{minipage}[t]{0.48\linewidth}
    \centering
    \includegraphics[width=\linewidth]{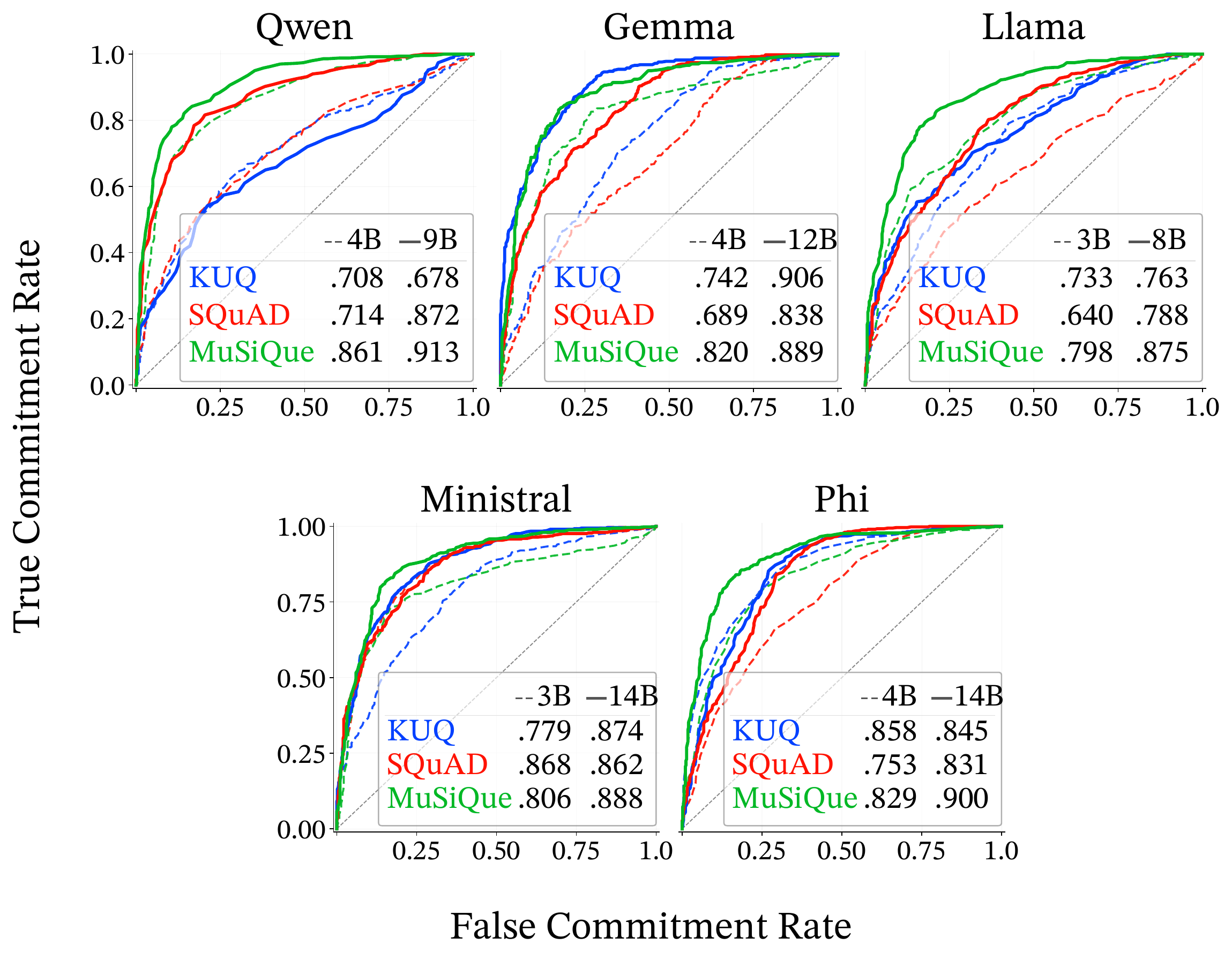}
    \caption{ROC curves for $\Delta(x)$ as a score for the answerable/unanswerable distinction across 30 configurations (AUROC in legend).}

    \label{fig:exp1-roc}
\end{minipage}%
\hfill
\begin{minipage}[t]{0.49\linewidth}
    \centering
    \includegraphics[width=\linewidth]{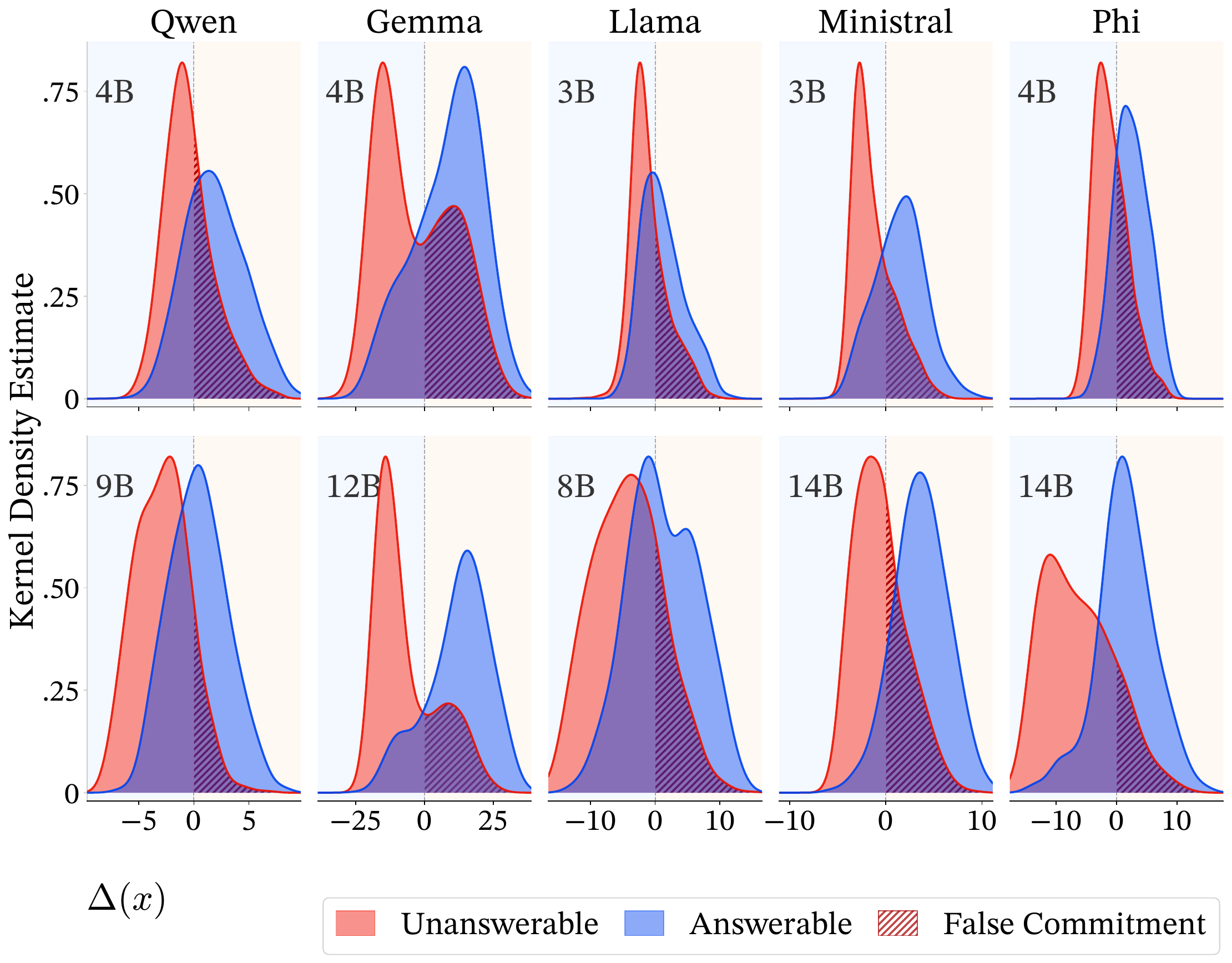}
    \caption{$\Delta(x)$ distributions  for answerable and unanswerable inputs. Hatched regions indicate FC, where $\Delta(x) \geq 0$ on unanswerable instances.}
    \label{fig:exp1-ridgeline}
\end{minipage}
\end{figure*}

We examine whether $\Delta(x)$ reflects signals of unanswerability above chance across model families and scales. For each instance $x$, we compute $\Delta(x)$ at the final input position. We evaluate $\Delta(x)$ as a score for the answerable/unanswerable distinction using AUROC, and report decision metrics under the threshold $\Delta(x) \geq 0$: decision accuracy (Acc), true commitment (TC), true abstention (TA), false abstention (FA), and false commitment (FC), our operational measure of unsupported commitment.

Figure~\ref{fig:exp1-roc} reports AUROC of $\Delta(x)$ across all $30$ model-dataset configurations, and Figure~\ref{fig:exp1-ridgeline} shows the distributions for answerable and unanswerable inputs. The sign of $\Delta(x)$ agrees with the model's actual commit-abstain behaviour from full generation in $97.6\%$ of instances on average (range $95.8\%$--$98.7\%$), supporting our formulation. Full per-configuration results are in App.~\ref{app:exp1-full}.

\textbf{Finding A: The margin $\Delta(x)$ reflects reliable signals of unanswerability, yet unsupported commitment persists.}
All $30$ AUROC values exceed chance, ranging from $0.640$ (Llama 3B on SQuAD~2.0) to $0.913$ (Qwen 9B on MuSiQue), with a mean of $0.811$. Within each model family, larger models generally achieve higher AUROC. The distribution of $\Delta(x)$ on unanswerable instances is consistently left-shifted, yet a substantial mass remains above zero in every configuration (hatched regions in Figure~\ref{fig:exp1-ridgeline}). On average, $27\%$ of unanswerable instances satisfy $\Delta(x) \geq 0$, ranging from $11\%$ (Qwen 9B) to $41\%$ (Gemma 4B). By Remark~1, since $\Delta(x)$ separates answerable from unanswerable inputs well above chance, the model's residual stream must carry information about unanswerability.

\subsection{Localising the CAC via Causal Component Gating}
\label{sec:gating-protocol}

\begin{figure*}[t]
\centering
\begin{minipage}[t]{0.65\textwidth}
  \centering
  \includegraphics[width=\textwidth]{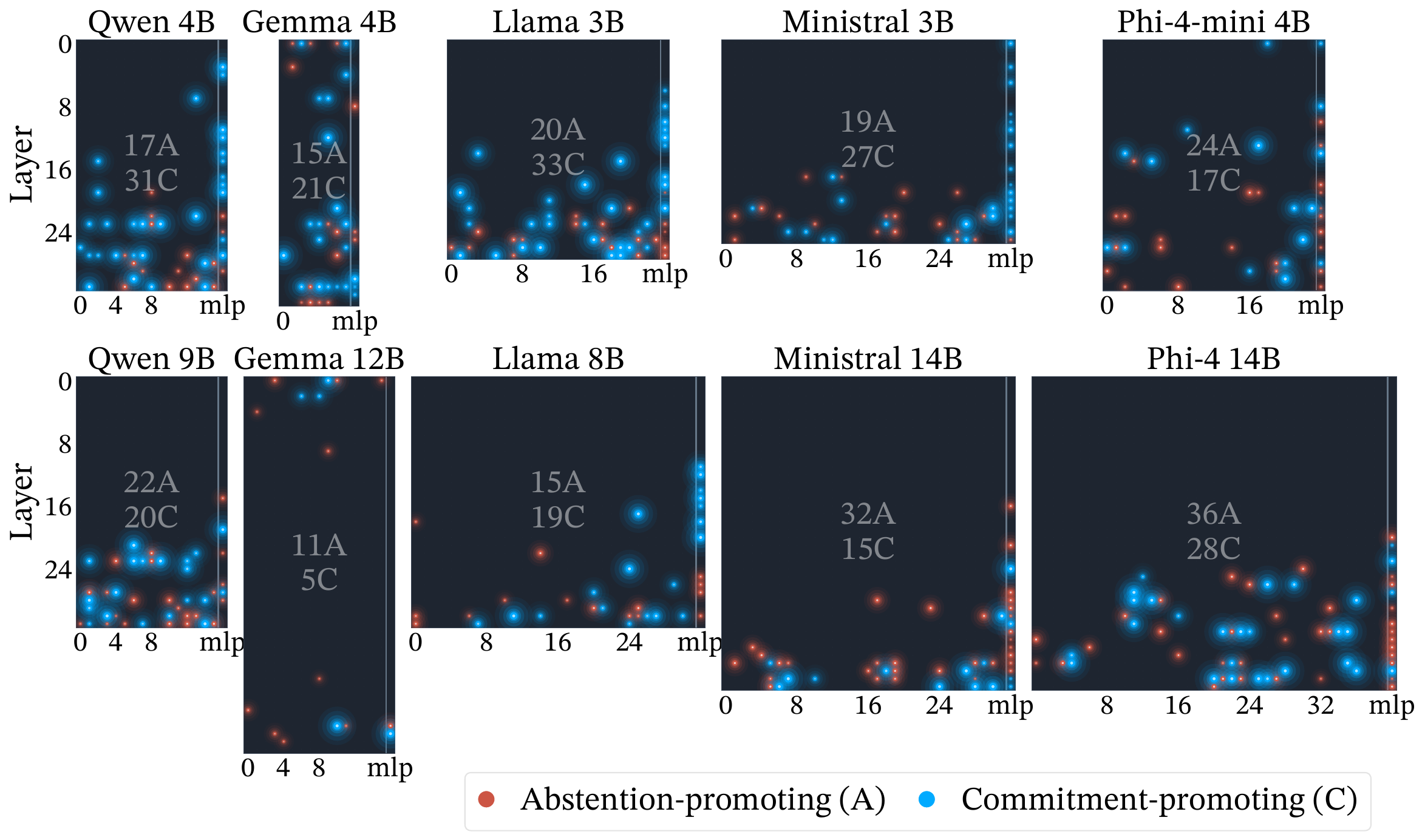}
    \caption{Identified CACs across ten models. Each panel plots layer (vertical) against head index with an additional MLP column (horizontal); dot brightness indicates mean causal strength across seeds. Inset numbers report the counts of abstention-promoting (A) and commitment-promoting (C) components per model.}
\label{fig:cac-circuit}
\end{minipage}
\hfill
\begin{minipage}[t]{0.33\textwidth}
  \centering
  \includegraphics[width=\linewidth]{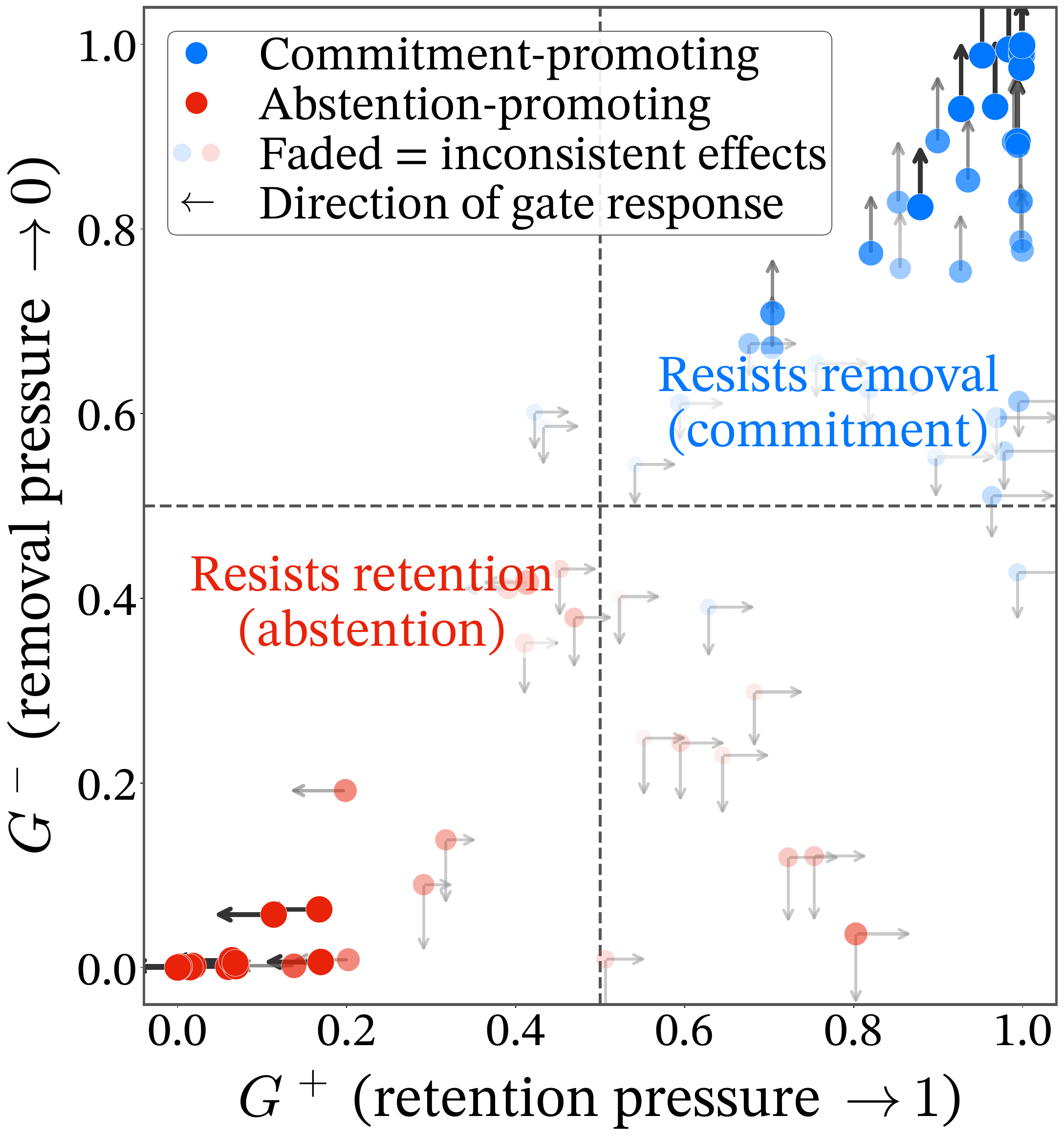}
  \caption{CAC identification via causal gating for Llama 3B. Each candidate component is plotted by its mean $G^+$ and $G^-$ values across $10$ seeds.}
  \label{fig:gate-scatter}
\end{minipage}
\end{figure*}

To study the commit-abstain decision internally, we identify the CAC via causal component gating. We adapt Causal Head Gating (CHG)~\citep{nam2025causal}, which assigns causal roles to attention heads under opposing regularisation pressures. Unlike causal mediation analysis~\citep{meng2022locating,XuCLLLW23,wang2023interpretability}, which typically intervenes on components separately, CHG optimises all gates jointly, allowing interacting components to be identified together~\citep{nam2025causal}. We extend CHG in two ways. First, we gate MLP sublayers alongside attention heads, motivated by prior work on MLP contributions to model behaviour~\citep{2025-causal-abstraction,dunefsky2024transcoders,chowdhury2026hedonic}. Second, we replace the next-token objective with a loss over the commit-abstain margin.

\paragraph{Gate Parameterisation}
For a transformer with $L$ layers and $H$ attention heads per layer, we assign a learnable logit $g \in \mathbb{R}$ to each attention head and MLP sublayer, with the corresponding gate defined as $G = \sigma(g)$, where $\sigma$ denotes the sigmoid function. The head gate scales the concatenated value vectors prior to the output projection $W^O_\ell$, while the MLP gate scales the sublayer output before residual addition. All base model weights are kept frozen; only the gate logits are optimised.

\paragraph{Task Loss}
Let $c^*(x)$ and $a^*(x)$ denote the top commit and abstain tokens under the ungated model, precomputed and held fixed so that gradients flow only through the gated logits. This fixing is a computational choice for gradient stability; the $97.6\%$ agreement between token-level and generation-level labels (Section~\ref{sec:unanswerability-signals}) indicates that this surrogate closely tracks commit-abstain behaviour. The task loss increases $\Delta(x)$ on answerable inputs and decreases it on unanswerable inputs:
\begin{equation}\label{eq:task-loss}
  \mathcal{L}_{\mathrm{task}}(G)
  = -\frac{1}{B} \sum_{x \in \mathrm{batch}} (2y - 1) \cdot \Delta(x \mid G),
  \qquad
  \Delta(x \mid G) = z_{c^*(x)}(G) - z_{a^*(x)}(G),
\end{equation}
where $y \in \{0,1\}$ is the answerability label, $(2y-1) \in \{-1,+1\}$ its sign, and $B$ is the batch size.

\paragraph{Gating Protocol}
Components that are important for the margin resist being turned off, whereas irrelevant components follow the applied regularisation pressure. Following \citet{nam2025causal}, we add opposing regularisation pressures to $\mathcal{L}_{\mathrm{task}}$:
\begin{equation}\label{eq:full-loss}
  \mathcal{L}(G;\lambda)
  = \mathcal{L}_{\mathrm{task}}(G)
  - \lambda \sum_{g \in \mathcal{S}}
    \mathrm{clip}(g,\,-C,\,C),
\end{equation}
where $\lambda$ is the regularisation strength, $C$ is the clipping bound, and $\mathcal{S}$ indexes the gated components.

We fit the gates in three phases:
\begin{itemize}
    \item \textbf{1. Task optimisation:} With $\lambda = 0$, we optimise the task loss to obtain an initialisation $\hat{G}$.

    \item \textbf{2a. Retention pressure:} With $\lambda > 0$, regularisation pushes gates towards $1$, yielding $G^+$.

    \item \textbf{2b. Removal pressure:} With $\lambda < 0$, regularisation pushes gates towards $0$, yielding $G^-$.
\end{itemize}

Phases~2a and 2b are both initialised from $\hat{G}$, so their differences reflect only the direction of regularisation.

The task-loss gradient on each gate logit is proportional to
$-(\mathbb{E}_{y=1}[v_k(x)]-\mathbb{E}_{y=0}[v_k(x)])$
(derivation in App.~\ref{app:gating-dynamics}). Thus, the loss tends to open gates for components that contribute more to $\Delta(x)$ on answerable inputs, and close gates for components that contribute more on unanswerable inputs. The resulting classification therefore reflects \emph{differential} contribution between the two subsets rather than the absolute sign of $v_k$.

A component is \textit{commitment-promoting} (\textsc{c}-component) if $G^- > \tau$, meaning that the task loss keeps it open despite removal pressure. A component is \textit{abstention-promoting} (\textsc{a}-component) if $G^+ < \tau$, meaning that the task loss keeps it closed despite retention pressure. We set $\tau = 0.5$, the sigmoid midpoint. Components meeting either criterion are classified as \textsc{a}-components if $\mathbb{E}[1-G^+] > \mathbb{E}[G^-]$, and as \textsc{c}-components otherwise; those meeting neither are excluded. Full hyperparameter details are provided in App.~\ref{app:gating-details}.

\subsection{Identifying the CAC}

We report the main findings here; full results, including screened candidates, per-model CACs, and cross-seed statistics, are provided in Appendix~\ref{app:exp2-full-results}.

\textbf{Finding B: Every tested model contains a sparse CAC.}
Figure~\ref{fig:cac-circuit} shows the identified CAC across all ten models. The CAC is sparse, covering $2.0\%$ (Gemma 12B) to $11.8\%$ (Gemma 4B) of components (median $5.2\%$). Both attention heads and MLP sublayers participate: on average, $25\%$ of CAC components are MLPs (range $12$--$34\%$), validating the extension beyond head-only gating. The three-phase gating protocol indicates that these components causally influence the commit-abstain margin. Components separate into three groups: those that remain closed under retention pressure (\textsc{a}-components), those that remain open under removal pressure (\textsc{c}-components), and those that follow the applied pressure (irrelevant). As shown in Figure~\ref{fig:gate-scatter} for Llama 3B, this separation is clear. Across models, $80\%$ of $G^+$ and $78\%$ of $G^-$ values lie below $0.1$ or above $0.9$, suggesting that component roles are resolved cleanly.

\begin{figure}[t]
\centering
\begin{minipage}[t]{0.31\columnwidth}
\centering
\vspace{0pt}
\includegraphics[width=1\columnwidth]{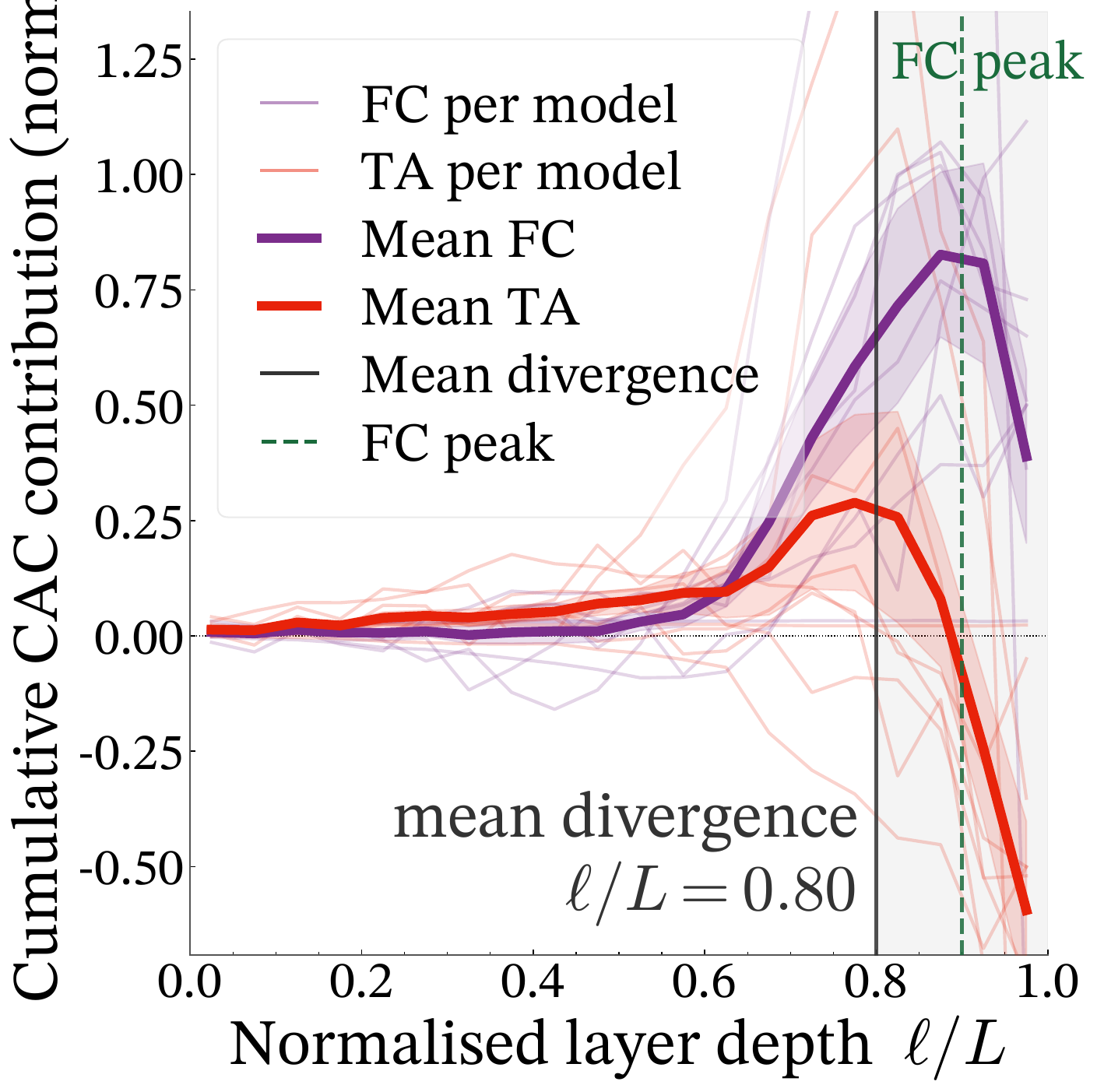}
\caption{Cumulative CAC contribution to $\Delta(x)$ by layer depth ($\ell/L$). FC and TA both accumulate positive contributions until 80\% depth, after which TA reverses sharply.}
\label{fig:trajectory}
\end{minipage}
\hfill
\begin{minipage}[t]{0.315\columnwidth}
\centering
\vspace{0pt}
\includegraphics[width=1\columnwidth]{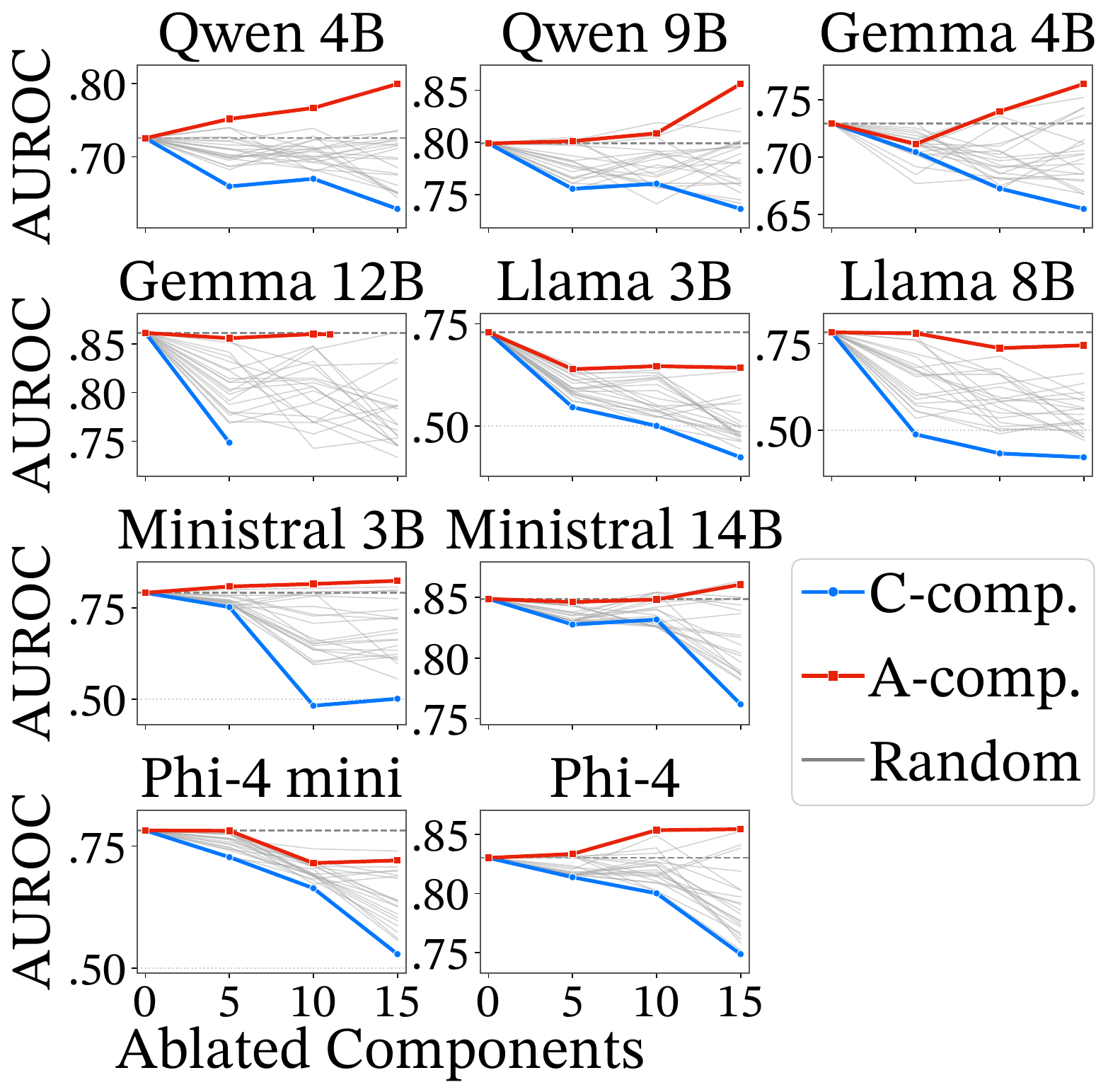}
\caption{Progressive ablation of \textsc{c}-components and \textsc{a}-components on held-out data against 20 random baselines (grey). Components are removed in order of causal score.}
\label{fig:ablation-auroc}
\end{minipage}
\hfill
\begin{minipage}[t]{0.31\columnwidth}
\centering
\vspace{0pt}
\includegraphics[width=1\columnwidth]{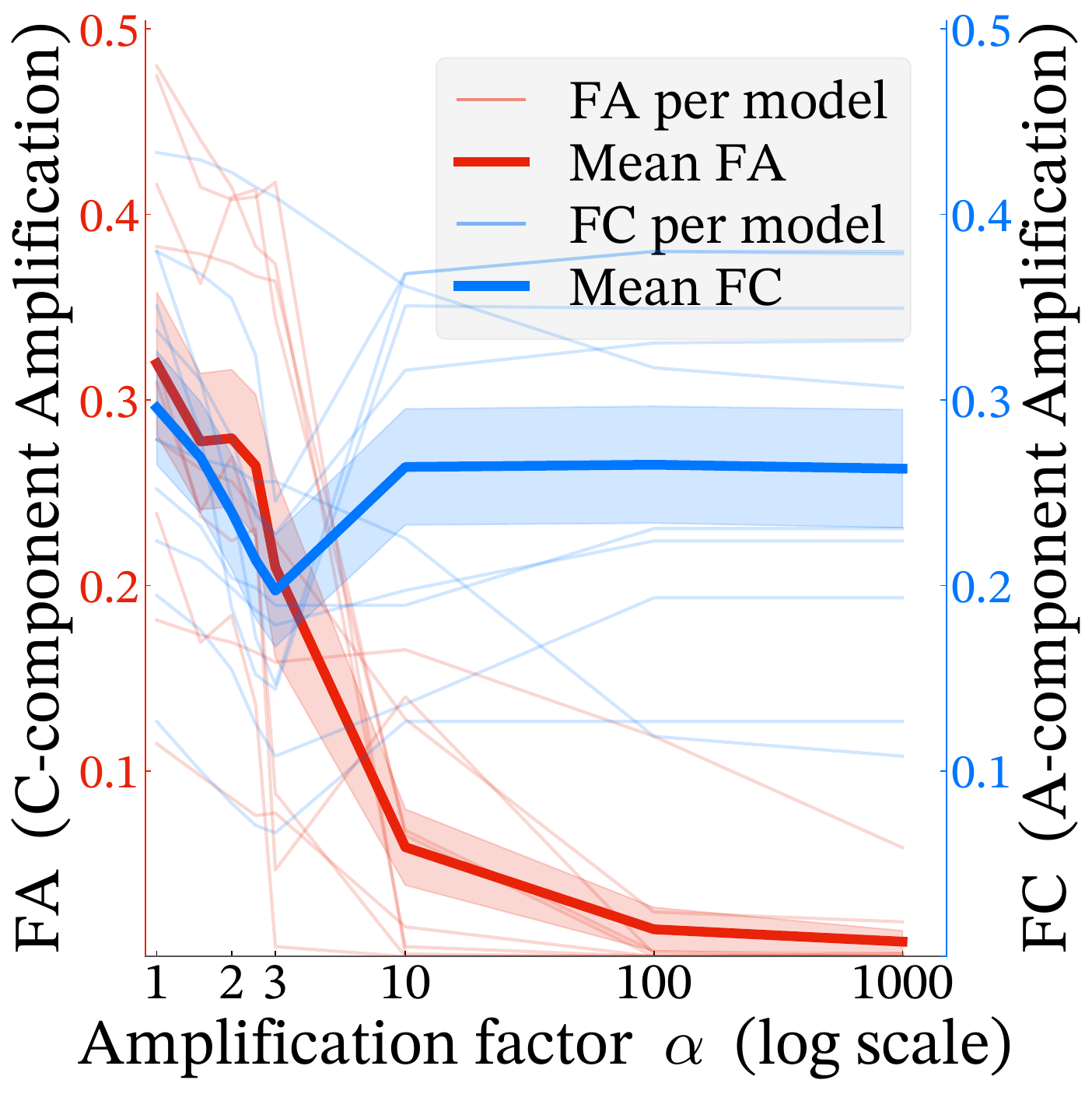}
\caption{\textsc{c}-component amplification drives FA to near zero by amplification factor $\alpha = 10$; \textsc{a}-component amplification leaves FC largely unchanged even at $\alpha = 1000$.}
\label{fig:asymmetry}
\end{minipage}
\hfill
\end{figure}

\textbf{Finding C: Commitment generally emerges in earlier layers than abstention.}
Whether the model ultimately commits or abstains, both FC and TA trajectories accumulate positive contributions until $80\%$ depth; only after this point does TA drop sharply negative while FC continues to rise, peaking at $90\%$ before also dropping (Figure~\ref{fig:trajectory}). At the component level, per \textsc{a}-component seen, more than $2\times$ as many \textsc{c}-components have appeared by $60\%$--$70\%$ depth in all models, $1.7\times$ at $80\%$, and only near parity by $90$--$100\%$ (App.~Table~\ref{tab:cac-summary}). Both results suggest that \textsc{c}-components are generally more concentrated in earlier layers than \textsc{a}-components.
The CAC is generally stable across random initialisations, screening thresholds, and regularisation strengths, and largely reproduces the sign of $\Delta(x)$, validating the identified components as robust contributors to the commit-abstain margin; full sensitivity analyses are provided in App.~\ref{app:sensitivity-analysis}.

\subsection{Functional Analysis of the CAC}
\label{sec:functional-analysis}
We verify the functional roles of the identified CAC components through two interventions: progressive ablation and targeted amplification. For ablation, we progressively remove the top-$k$ components on the test set, ordered by gating score, controlling against 20 size-matched random non-CAC ablations to verify that observed effects are CAC-specific. For amplification, we scale each component output direction $h$ by $\alpha \in \{1.0, 1.5, 2.0, 2.5, 3.0, 10, 100, 1000\}$ before residual addition to test whether each type causally contributes to FA and FC rates. Full results are in App.~\ref{app:exp3-full}.

\textbf{Finding D: \textsc{c}-components sustain the discriminative structure of $\Delta(x)$; \textsc{a}-components act correctionally but with insufficient intrinsic strength to overcome accumulated commitment.} Ablating \textsc{a}-components causes $17.1\%$ abstain-to-commit (A$\to$C) flips versus $7.8\%$ commit-to-abstain (C$\to$A), yet AUROC remains stable (mean $-0.03$) as flipped inputs are borderline, shifting from mildly abstaining ($\bar\Delta = {-}2.78$) to committing ($\bar\Delta = {+}3.99$) with rank largely preserved. Ablating \textsc{c}-components, counterintuitively, also yields A$\to$C-skewed flips ($31.7\%$ vs.\ $3.4\%$). AUROC drops in every model (mean $-0.17$; worst: Qwen~4B $0.73 \to 0.40$), continuing to decline as more components are removed while size-matched random ablation changes it by only $-0.069$ (Figure~\ref{fig:ablation-auroc}). Both ablations skew A$\to$C, but for distinct reasons: \textsc{a}-components push $\Delta(x)$ correctionally across 0 with little effect on ranking structure; \textsc{c}-components sustain the discriminative structure of $\Delta(x)$, their removal collapsing the separation between answerable and unanswerable inputs globally.

Amplification further validates these roles. At moderate $\alpha$ (${\leq}3$), \textsc{c}-component amplification shifts decisions toward commitment, reducing FA from $0.320$ to $0.210$ while increasing FC from $0.296$ to $0.418$. Conversely, \textsc{a}-component amplification shifts decisions toward abstention, reducing FC from $0.296$ to $0.197$ while increasing FA from $0.320$ to $0.415$. This pattern holds across all $10$ models, with mean FA and FC reductions of $37.6\%$ and $33.9\%$, respectively. At extreme $\alpha$ (${\geq}10$), the two sweeps diverge sharply: FA drops to near zero under \textsc{c}-component amplification, whereas FC stabilises well above zero under \textsc{a}-component amplification, with FC at $\alpha{=}100$ nearly identical to $\alpha{=}1000$ (Figure~\ref{fig:asymmetry}).

This divergence can be understood through the behaviour of Root Mean Square Normalisation (RMSNorm) under large scaling:
\begin{equation*}
\text{RMSNorm}(R + \alpha h)
= \frac{R + \alpha h}{\text{RMS}(R + \alpha h)} \cdot \gamma
\approx
\frac{\alpha h}{\text{RMS}(\alpha h)} \cdot \gamma
=
\frac{h}{\text{RMS}(h)} \cdot \gamma
= \hat{h},
\quad
\text{when } \alpha \gg \frac{\|R\|}{\|h\|},
\end{equation*}
where $R$ is the residual stream and $\gamma$ is a learned vector fixed at inference time. Intuitively, when $\alpha$ is sufficiently large, both $\alpha$ and the pre-existing residual $R$ cease to dominate the normalised representation: the next layer is driven primarily by the component direction $\hat{h}$. Extreme amplification therefore probes the component's intrinsic downstream effect with prior residual accumulation largely normalised out.

The resulting asymmetry is informative. A \textsc{c}-component direction reliably drives $\Delta(x)$ upward and can nearly eliminate false abstention under strong amplification. In contrast, an \textsc{a}-component direction shifts decisions toward abstention at moderate amplification but cannot eliminate false commitment even under extreme scaling. Thus, \textsc{c}-components provide a strong commitment-driving signal, whereas \textsc{a}-components act as corrective signals that are not independently sufficient to guarantee abstention.

\paragraph{Remark 2}
The model encodes unanswerability internally (Finding~A) and contains a sparse CAC (Finding~B), yet may still commit because abstention-promoting contributions arise later (Finding~C) and are often insufficient to drive $\Delta(x)$ below zero (Finding~D). These findings suggest a component-level account of unsupported commitment: the model commits not because abstention-related signals are absent, but because their corrective effects fail to overturn accumulated commitment.

\begin{table}[t]
\centering
\caption{Mean AUROC and accuracy across three datasets for each model. \textbf{Bold} indicates the best result; \underline{underline} indicates the second best. $p$-values are from two-sided Wilcoxon signed-rank tests over $30$ model-dataset pairs. All differences are significant ($p < 0.05$), with most at $p < 10^{-5}$.}
\label{tab:main-results}
\renewcommand{\arraystretch}{1}
\setlength{\tabcolsep}{4.5pt}
\footnotesize
\begin{tabular}{c|l cccccccccc c}
\toprule
\multicolumn{2}{c}{} & \multicolumn{2}{c}{Qwen 3.5} & \multicolumn{2}{c}{Gemma 3} & \multicolumn{2}{c}{Llama 3} & \multicolumn{2}{c}{Ministral} & \multicolumn{2}{c}{Phi-4} & \\
\cmidrule(lr){3-4} \cmidrule(lr){5-6} \cmidrule(lr){7-8} \cmidrule(lr){9-10} \cmidrule(lr){11-12}
\multicolumn{2}{c}{} & 4B & 9B & 4B & 12B & 3B & 8B & 3B & 14B & mini & 14B & $p$ \\
\midrule
\multirow{9}{*}{\rotatebox{90}{Mean AUROC}}
& Zero-Threshold & \cellcolor{gray!17}.725 & \cellcolor{gray!26}.798 & \cellcolor{gray!17}.722 & \cellcolor{gray!35}\underline{.854} & \cellcolor{gray!14}.689 & \cellcolor{gray!24}.784 & \cellcolor{gray!25}.790 & \cellcolor{gray!34}.848 & \cellcolor{gray!24}.780 & \cellcolor{gray!31}.830 & $6.0{\times}10^{-6}$ \\
& Non-CAC Random & \cellcolor{gray!14}.691 & \cellcolor{gray!22}.764 & \cellcolor{gray!19}.741 & \cellcolor{gray!27}.808 & \cellcolor{gray!11}.648 & \cellcolor{gray!28}.815 & \cellcolor{gray!18}.732 & \cellcolor{gray!25}.789 & \cellcolor{gray!21}.756 & \cellcolor{gray!30}.822 & $8.4{\times}10^{-6}$ \\
& INSIDE & \cellcolor{gray!5}.572 & \cellcolor{gray!1}.449 & \cellcolor{gray!6}.586 & \cellcolor{gray!11}.658 & \cellcolor{gray!6}.578 & \cellcolor{gray!2}.483 & \cellcolor{gray!1}.406 & \cellcolor{gray!2}.493 & \cellcolor{gray!1}.415 & \cellcolor{gray!2}.492 & $3.7{\times}10^{-9}$ \\
& Multi-LLM & \cellcolor{gray!13}.688 & \cellcolor{gray!21}.757 & \cellcolor{gray!13}.683 & \cellcolor{gray!18}.728 & \cellcolor{gray!4}.538 & \cellcolor{gray!15}.705 & \cellcolor{gray!7}.596 & \cellcolor{gray!11}.654 & \cellcolor{gray!7}.603 & \cellcolor{gray!15}.703 & $1.9{\times}10^{-9}$ \\
& Semantic Entropy & \cellcolor{gray!5}.567 & \cellcolor{gray!2}.463 & \cellcolor{gray!4}.548 & \cellcolor{gray!7}.600 & \cellcolor{gray!5}.564 & \cellcolor{gray!3}.521 & \cellcolor{gray!1}.423 & \cellcolor{gray!5}.564 & \cellcolor{gray!3}.509 & \cellcolor{gray!4}.529 & $1.3{\times}10^{-8}$ \\
& MERA & \cellcolor{gray!12}.675 & \cellcolor{gray!5}.574 & \cellcolor{gray!9}.635 & \cellcolor{gray!11}.657 & \cellcolor{gray!8}.613 & \cellcolor{gray!5}.560 & \cellcolor{gray!10}.648 & \cellcolor{gray!20}.747 & \cellcolor{gray!9}.634 & \cellcolor{gray!4}.532 & $1.9{\times}10^{-9}$ \\
& HaloScope & \cellcolor{gray!14}.697 & \cellcolor{gray!9}.627 & \cellcolor{gray!17}.719 & \cellcolor{gray!14}.693 & \cellcolor{gray!24}\underline{.780} & \cellcolor{gray!19}.744 & \cellcolor{gray!11}.665 & \cellcolor{gray!26}.792 & \cellcolor{gray!14}.689 & \cellcolor{gray!15}.700 & $1.9{\times}10^{-8}$ \\
& HaMI & \cellcolor{gray!39}\underline{.876} & \cellcolor{gray!48}\textbf{.919} & \cellcolor{gray!21}\underline{.761} & \cellcolor{gray!32}.833 & \cellcolor{gray!21}.754 & \cellcolor{gray!34}\underline{.845} & \cellcolor{gray!27}\underline{.802} & \cellcolor{gray!39}\underline{.874} & \cellcolor{gray!33}\textbf{.838} & \cellcolor{gray!40}\underline{.878} & $2.0{\times}10^{-3}$ \\
& CAC-Informed MLP (Ours) & \cellcolor{gray!40}\textbf{.879} & \cellcolor{gray!46}\underline{.911} & \cellcolor{gray!28}\textbf{.807} & \cellcolor{gray!42}\textbf{.889} & \cellcolor{gray!28}\textbf{.807} & \cellcolor{gray!39}\textbf{.874} & \cellcolor{gray!39}\textbf{.874} & \cellcolor{gray!39}\textbf{.877} & \cellcolor{gray!32}\underline{.837} & \cellcolor{gray!42}\textbf{.888} & -- \\
\midrule
\multirow{9}{*}{\rotatebox{90}{Mean Accuracy}}
& Zero-Threshold & \cellcolor{gray!11}.655 & \cellcolor{gray!14}.697 & \cellcolor{gray!10}.644 & \cellcolor{gray!23}.770 & \cellcolor{gray!8}.623 & \cellcolor{gray!10}.651 & \cellcolor{gray!16}.714 & \cellcolor{gray!20}.753 & \cellcolor{gray!15}.705 & \cellcolor{gray!16}.711 & $2.6{\times}10^{-6}$ \\
& Non-CAC Random & \cellcolor{gray!9}.628 & \cellcolor{gray!17}.724 & \cellcolor{gray!13}.682 & \cellcolor{gray!19}.744 & \cellcolor{gray!7}.596 & \cellcolor{gray!18}.731 & \cellcolor{gray!12}.670 & \cellcolor{gray!16}.718 & \cellcolor{gray!14}.693 & \cellcolor{gray!20}.752 & $4.2{\times}10^{-6}$ \\
& INSIDE & \cellcolor{gray!5}.571 & \cellcolor{gray!3}.523 & \cellcolor{gray!7}.601 & \cellcolor{gray!15}.708 & \cellcolor{gray!5}.564 & \cellcolor{gray!4}.538 & \cellcolor{gray!3}.515 & \cellcolor{gray!4}.539 & \cellcolor{gray!3}.515 & \cellcolor{gray!4}.543 & $1.9{\times}10^{-6}$ \\
& Multi-LLM & \cellcolor{gray!12}.673 & \cellcolor{gray!14}.691 & \cellcolor{gray!12}.673 & \cellcolor{gray!17}.727 & \cellcolor{gray!7}.597 & \cellcolor{gray!14}.693 & \cellcolor{gray!11}.655 & \cellcolor{gray!13}.678 & \cellcolor{gray!9}.632 & \cellcolor{gray!11}.661 & $1.7{\times}10^{-6}$ \\
& Semantic Entropy & \cellcolor{gray!6}.576 & \cellcolor{gray!3}.506 & \cellcolor{gray!5}.561 & \cellcolor{gray!7}.602 & \cellcolor{gray!5}.555 & \cellcolor{gray!4}.539 & \cellcolor{gray!3}.500 & \cellcolor{gray!5}.558 & \cellcolor{gray!4}.537 & \cellcolor{gray!5}.555 & $1.9{\times}10^{-6}$ \\
& MERA & \cellcolor{gray!10}.640 & \cellcolor{gray!6}.592 & \cellcolor{gray!6}.588 & \cellcolor{gray!10}.646 & \cellcolor{gray!5}.573 & \cellcolor{gray!6}.580 & \cellcolor{gray!8}.621 & \cellcolor{gray!11}.665 & \cellcolor{gray!7}.597 & \cellcolor{gray!5}.565 & $1.9{\times}10^{-9}$ \\
& HaloScope & \cellcolor{gray!7}.595 & \cellcolor{gray!11}.655 & \cellcolor{gray!13}.685 & \cellcolor{gray!11}.653 & \cellcolor{gray!15}\underline{.708} & \cellcolor{gray!16}.715 & \cellcolor{gray!11}.663 & \cellcolor{gray!14}.691 & \cellcolor{gray!14}.691 & \cellcolor{gray!12}.671 & $1.7{\times}10^{-6}$ \\
& HaMI & \cellcolor{gray!30}\underline{.822} & \cellcolor{gray!40}\textbf{.881} & \cellcolor{gray!16} \underline{.711} & \cellcolor{gray!25}\underline{.785} & \cellcolor{gray!15}.707 & \cellcolor{gray!27}\underline{.803} & \cellcolor{gray!19}\underline{.740} & \cellcolor{gray!34}\textbf{.849} & \cellcolor{gray!25}\textbf{.786} & \cellcolor{gray!31}\underline{.831} & $1.6{\times}10^{-2}$ \\
& CAC-Informed MLP (Ours) & \cellcolor{gray!31}\textbf{.829} & \cellcolor{gray!35}\underline{.855} & \cellcolor{gray!22}\textbf{.765} & \cellcolor{gray!34}\textbf{.845} & \cellcolor{gray!21}\textbf{.755} & \cellcolor{gray!30}\textbf{.820} & \cellcolor{gray!29}\textbf{.815} & \cellcolor{gray!31}\underline{.829} & \cellcolor{gray!25}\underline{.785} & \cellcolor{gray!33}\textbf{.840} & -- \\
\bottomrule
\end{tabular}

\end{table}

\section{Learning an Abstention Policy from the CAC}
\label{sec:policy-cac}
We now address RQ2 by testing whether the CAC is actionable beyond characterising the commit-abstain decision. To this end, we evaluate a lightweight abstention policy trained on CAC activations.

\subsection{CAC-Informed Abstention Policy}

For each input $x$, we extract the  contribution $v_k(x)$ of each CAC component to $\Delta(x)$ at the final input position, and append aggregate statistics over the CAC vector (mean, standard deviation, min, max, and summed contributions of \textsc{a}- and \textsc{c}-components). Concatenating yields a $d_{\text{in}}$-dimensional feature vector, where $d_{\text{in}}$ is $16$--$64$ plus statistics.

We train a lightweight MLP with architecture $d_{\text{in}} \to 32 \to 128 \to 64 \to 32 \to 1$. The initial bottleneck projects variable-dimensional CAC features into a fixed $32$-dimensional representation before non-linear feature mixing, providing dimensionality normalisation and regularisation. All hidden layers use batch normalisation, ReLU activations, and dropout ($p{=}0.15$). We train with binary cross-entropy using AdamW ($\text{lr}{=}10^{-3}$, $\text{wd}{=}10^{-3}$) for up to $500$ epochs with early stopping (patience $120$). At inference, the model outputs $P(\text{answerable})$, and abstention is triggered when this probability falls below a threshold calibrated on a validation set held out from the training split.
\subsection{Experiments \& Results}

\paragraph{Baselines}
We compare against \textbf{Zero-Threshold}, which abstains when $\Delta(x) < 0$; a size-matched \textbf{Non-CAC Random} MLP that controls for whether the gain is CAC-specific; and six established methods. \textbf{INSIDE}~\citep{chen2024inside} computes EigenScores from the covariance of multiple sampled response embeddings. \textbf{Multi-LLM}~\citep{feng-etal-2024-dont} abstains based on cross-LLM agreement. \textbf{Semantic Entropy}~\citep{Farquhar2024} measures entropy over semantically clustered generations. \textbf{MERA}~\citep{hedstrom2025to} is an activation-steering framework with calibrated abstention. \textbf{HaloScope}~\citep{du2024haloscope} trains a binary classifier on a hallucination subspace identified via SVD on hidden states. \textbf{HaMI}~\citep{niu2025robust} is an MLP detector trained via Multiple Instance Learning over adaptively-selected token activations. While we also use an MLP, our method differs from HaMI in feature space ($16$--$64$ CAC components vs.\ full $d_{\text{model}}$-dimensional activations) and selection principle (localisation of model components vs.\ selection of hallucination-indicative tokens). For comparability, all baselines are trained or calibrated using the same training split as our method (\textbf{CAC-Informed MLP}). Detailed configurations are in App.~\ref{app:baselines}.

\paragraph{Main Results}
Table~\ref{tab:main-results} reports the main results on the test split, with per-dataset metrics in App.~\ref{app:exp4-full-results}. Our policy achieves the highest mean AUROC ($0.864$) and decision accuracy ($0.814$) across all $30$ configurations, with substantial gains over the Zero-Threshold baseline ($+8.2$ AUROC points, $+12.2$ accuracy points). It also outperforms HaMI on $23$ of $30$ configurations in AUROC and $20$ of $30$ in accuracy. Notably, these improvements do not come at the cost of over-abstention: the mean false abstention rate is reduced from $0.320$ to $0.127$ ($2.5\times$ lower). These results suggest that the per-component CAC features carry richer information prior to additive aggregation in the residual stream: while $\Delta(x)$ collapses this aggregation into a scalar, the policy retains the per-component contributions and models nonlinear interactions across components.

\paragraph{Out-of-Distribution (OOD) Generalisation} We evaluate generalisation along two axes: (i) extension to larger models (Gemma-3 27B and Qwen-3.5 35B), and (ii) transfer to unseen datasets, HotpotQA~\citep{yang-etal-2018-hotpotqa} and SelfAware~\citep{yin-etal-2023-large}. For dataset transfer, the trained policy is applied without fine-tuning. For larger models, we localise the CAC and train the same MLP architecture on the in-distribution datasets before evaluating on unseen datasets. Table~\ref{tab:generalisation-accuracy} reports results across all $12$ models and both unseen datasets. The CAC-informed policy improves over Zero-Threshold in all $24$ OOD configurations, with mean gains of $+6.9$ accuracy points. On the two larger models, the policy gains $+6.3$ accuracy points on average. Notably, it also outperforms the strongest baseline HaMI in $19$ of $24$ configurations. These results suggest that the localisation pipeline reproduces at larger scales, and that the policy generalises zero-shot across datasets.

\begin{table}[t]
    \centering
    \caption{%
        Decision accuracy on two unseen datasets (HotpotQA, SelfAware) and two unseen larger models ({$^\dagger$}). Our method achieves higher accuracy in all $24$ configurations ($p = 1.8{\times}10^{-5}$).
    }
    \label{tab:generalisation-accuracy}
    \renewcommand{\arraystretch}{1}
    \setlength{\tabcolsep}{4.8pt}
    \footnotesize
    \begin{tabular}{l cccccccccccc}
    \toprule
    & \multicolumn{3}{c}{Qwen 3.5} & \multicolumn{3}{c}{Gemma 3} & \multicolumn{2}{c}{Llama 3} & \multicolumn{2}{c}{Ministral} & \multicolumn{2}{c}{Phi-4} \\
    \cmidrule(lr){2-4} \cmidrule(lr){5-7} \cmidrule(lr){8-9} \cmidrule(lr){10-11} \cmidrule(lr){12-13}
     & 4B & 9B & {35B$^\dagger$} & 4B & 12B & {27B$^\dagger$} & 3B & 8B & 3B & 14B & mini & 14B \\
    \midrule
    \multicolumn{13}{c}{\textit{HotpotQA}} \\ \midrule
    Zero-Threshold & .887 & .837 & {.867} & \underline{.898} & \underline{.922} & {\underline{.924}} & \underline{.840} & .820 & .851 & \underline{.892} & \underline{.859} & .842 \\
    HaMI & \underline{.917} & \underline{.938} & {\underline{.925}} & .894 & .915 & {.892} & .829 & \textbf{.924} & \underline{.865} & .889 & .856 & \underline{.937} \\
    CAC-Informed MLP (Ours) & \textbf{.930} & \textbf{.949} & {\textbf{.928}} & \textbf{.911} & \textbf{.948} & {\textbf{.950}} & \textbf{.859} & \underline{.913} & \textbf{.895} & \textbf{.894} & \textbf{.875} & \textbf{.939} \\
    \midrule
    \multicolumn{13}{c}{\textit{SelfAware}} \\ \midrule
    Zero-Threshold & .639 & .544 & {.620} & .637 & .711 & {\underline{.747}} & .660 & .655 & .698 & .709 & \underline{.717} & .697 \\
    HaMI & \textbf{.761} & \textbf{.749} & {\underline{.649}} & \underline{.682} & \underline{.755} & {.734} & \underline{.675} & \textbf{.782} & \textbf{.773} & \underline{.738} & .713 & \underline{.767} \\
    CAC-Informed MLP (Ours) & \underline{.758} & \underline{.746} & {\textbf{.773}} & \textbf{.737} & \textbf{.763} & {\textbf{.757}} & \textbf{.734} & \underline{.764} & \underline{.769} & \textbf{.799} & \textbf{.748} & \textbf{.779} \\
    \bottomrule
    \end{tabular}
    
    \end{table}

We evaluate a logistic regressor over the same features, achieving $+10.6$ accuracy points over Zero-Threshold (vs.\ $+12.2$ for the MLP, App.~\ref{app:regression}), indicating that much of the gain comes from directly reading CAC features, with the MLP capturing additional nonlinear interactions across components. We include two case studies of unsupported commitment in App.~\ref{app:case-studies}.

\section{Conclusion}

Unsupported commitment reflects a representation-behaviour gap: models carry signals of unanswerability internally yet still commit. By formalising the commit-abstain margin and applying causal gating, we identified a sparse Commit-Abstain Circuit in every tested model and uncovered a recurring pattern: commitment-promoting components act earlier and accumulate momentum, while abstention-promoting components arise later as corrective signals that often fail to overturn what has already been built up. This provides an intuitive mechanistic account of how unsupported commitment can emerge even when abstention-related signals are present internally. More broadly, this suggests new directions for mitigation. Rather than only detecting hallucinations after they emerge or training models to abstain more often, internal mechanistic signals may enable earlier intervention and more targeted inference-time control over when the model should commit or defer. In this sense, the CAC offers not only an explanation of unsupported commitment, but also a foundation for more reliable model behaviour.

\paragraph{Limitations}
Several limitations should be considered when interpreting these findings. First, our scope is restricted to unsupported commitment on unanswerable inputs in instruction-tuned models; we do not address incorrect answers to otherwise answerable questions. Second, our main analysis uses the pre-generation commit-abstain margin. Although this agrees with full-generation behaviour in $97.6\%$ of cases and CAC signals track observed mid-generation reversals, generation-time dynamics remain only partially characterised. Third, both the abstention readout and the CAC provide partial operationalisations of the underlying behaviour. Recent work has emphasised falsification-based evaluation to distinguish genuine behavioural mechanisms from superficial lexical correlates~\citep{ma2026do}. In our setting, $\mathcal{A}$ captures common lexical forms of epistemic deferral but may not cover all abstention behaviours, while the CAC is a causally localised, group-level account rather than a complete functional characterisation of every component or interaction. Future work should further stress-test these claims across broader behavioural settings.

\section*{Acknowledgments}
This research was supported in part by the Australian Research Council Discovery Project DP260104188 and the RMIT University School of Computing Technologies Research Support Package. Renqiang Luo was supported by the GBA Ascend Application Innovation Institute (GML-ST-2026-24), Guangdong Laboratory of Artificial Intelligence and Digital Economy (SZ).

\bibliographystyle{unsrtnat}
\bibliography{references}
\newpage


\appendix

\section{Datasets}
\label{app:datasets}

The five datasets are chosen to stress-test abstention across three orthogonal axes of difficulty: \textit{knowledge source} (parametric recall vs.\ in-context retrieval), \textit{reasoning depth} (single-hop vs.\ multi-hop), and \textit{unanswerability mechanism}.
Table~\ref{tab:datasets} summarises their key properties.

\begin{table}[h]
\centering
\caption{%
  Dataset summary. \emph{Source}: whether the model must rely on parametric knowledge or a provided context passage. ($\dagger$) held-out OOD evaluation sets only.
}
\label{tab:datasets}
\setlength{\tabcolsep}{4pt}
\footnotesize
\begin{tabular*}{\textwidth}{@{\extracolsep{\fill}} l l p{0.15\textwidth} p{0.42\textwidth} l@{}}
\toprule
\textbf{Dataset} & \textbf{Source} & \textbf{Reasoning} & \textbf{Unanswerability} & \textbf{License} \\
\midrule
\textbf{KUQ}~\citep{amayuelas-etal-2024-knowledge}
  & Parametric
  & Open-domain QA
  & Answers genuinely do not exist (unsolved problems, future events); tests limits of \textit{human} knowledge.
  & MIT \\
\cmidrule(r){1-1}\cmidrule(lr){2-2}\cmidrule(lr){3-3}\cmidrule(lr){4-4}\cmidrule(l){5-5}
\textbf{SQuAD 2.0}~\citep{rajpurkar-etal-2018-know}
  & In-context
  & Reading comprehension
  & Questions plausible but unsupported by the passage; tests resistance to surface-level cues.
  & CC BY-SA 4.0 \\
\cmidrule(r){1-1}\cmidrule(lr){2-2}\cmidrule(lr){3-3}\cmidrule(lr){4-4}\cmidrule(l){5-5}
\textbf{MuSiQue}~\citep{trivedi-etal-2022-musique}
  & In-context
  & Multi-hop
  & One supporting document removed (\textit{missing evidence}); tests detection of a silent gap in the reasoning chain.
  & CC BY 4.0 \\
\midrule
\textbf{SelfAware}$^\dagger$~\citep{yin-etal-2023-large}
  & Parametric
  & Self-knowledge
  & No definitive answer exists; tests limits of the model's \textit{own} knowledge, not of the world.
  & CC BY-SA 4.0 \\
\cmidrule(r){1-1}\cmidrule(lr){2-2}\cmidrule(lr){3-3}\cmidrule(lr){4-4}\cmidrule(l){5-5}
\textbf{HotpotQA}$^\dagger$~\citep{yang-etal-2018-hotpotqa}
  & In-context
  & Multi-hop
  & Supporting passages replaced by distractors (\textit{misleading evidence}); tests resistance to plausible-but-wrong context.
  & CC BY-SA 4.0 \\
\bottomrule
\end{tabular*}
\end{table}

\section{Construction and Validation of the Abstention Token Set \texorpdfstring{$\mathcal{A}$}{A}}
\label{app:token-families}
Table~\ref{tab:token-families} lists the $17$ tokens in $\mathcal{A}$.
We built $\mathcal{A}$ using an additional pool of $1{,}000$ unanswerable instances drawn from the selected datasets, disjoint from the train and test splits, so the abstention vocabulary is constructed independently of the instances used for CAC localisation and evaluation. We ran zero-shot inference on this held-aside pool across all ten models, recorded the first generated token matching a canonical deferral pattern, then pooled and deduplicated across models. We restrict $\mathcal{A}$ to tokens that unambiguously initiate epistemic deferral, so the partition $(\mathcal{A}, \mathcal{C})$ is fixed at evaluation time and $\Delta(x)$ is a deterministic function of $r_L(x)$.

\begin{figure}[t]
\centering
\includegraphics[width=0.98\textwidth]{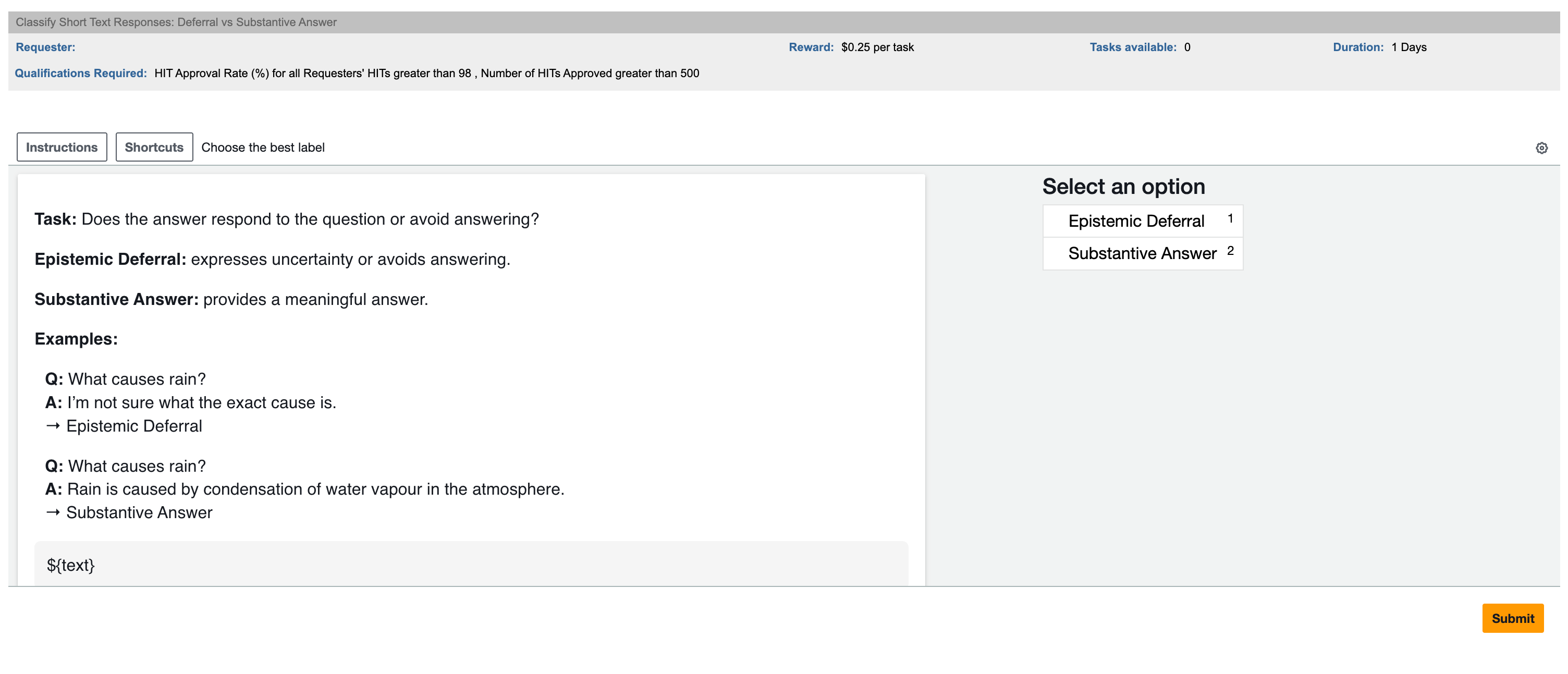}
\caption{%
Annotation interface shown to Amazon Mechanical Turk workers for classifying candidate tokens as epistemic deferral or substantive answer.}
\label{fig:annotation-instructions}
\end{figure}

Each candidate token with its $10$-token continuation was presented to two annotators on Amazon Mechanical Turk (English-speaking, $\geq 98\%$ approval rate, $\geq 500$ approved HITs), who classified each item as \emph{epistemic deferral} (abstention) or \emph{substantive answer} (commitment); the annotation interface and instructions are shown in Fig.~\ref{fig:annotation-instructions}. Inter-annotator agreement was $98.3\%$ (Cohen's $\kappa = 0.96$); disagreements were resolved by majority vote among three additional annotators. Workers were compensated at \$20/hour.

\begin{table}[t]
\centering
\caption{%
  Abstention token set $\mathcal{A}$ ($17$ tokens).
  The commitment set $\mathcal{C}$ contains all tokens not in $\mathcal{A}$.
}
\label{tab:token-families}
\setlength{\tabcolsep}{3pt}
\footnotesize
\begin{tabular}{@{} l @{\hspace{4pt}} p{0.34\textwidth} @{\hspace{10pt}} l @{\hspace{4pt}} p{0.34\textwidth} @{}}
\toprule
\texttt{I}             & ``I don't know'', ``I cannot determine'' & \texttt{Sorry}         & ``Sorry, I don't have information\ldots'' \\
\texttt{Unfortunately} & ``Unfortunately, I don't have enough\ldots'' & \texttt{Beyond}        & ``Beyond my knowledge\ldots'' \\
\texttt{Insufficient}  & ``Insufficient information to answer\ldots'' & \texttt{Impossible}    & ``Impossible to determine without context'' \\
\texttt{Unanswerable}  & ``Unanswerable based on context'' & \texttt{Without}       & ``Without more context, I cannot\ldots'' \\
\texttt{Unclear}       & ``Unclear from the given context\ldots'' & \texttt{Ambiguous}     & ``Ambiguous; the premise is unclear'' \\
\texttt{Unknown}       & ``Unknown; context does not specify\ldots'' & \texttt{Unsure}        & ``Unsure; evidence is inconclusive'' \\
\texttt{Uncertain}     & ``Uncertain; lacks a definitive answer'' & \texttt{Unknowable}    & ``Unknowable from the text provided'' \\
\texttt{Cannot}        & ``Cannot be determined from context'' & \texttt{Indeterminate} & ``Indeterminate based on context'' \\
\texttt{Unable}        & ``Unable to answer from the passage'' & & \\
\bottomrule
\end{tabular}
\end{table}

\paragraph{Validation: $\Delta(x)$ agreement with full-generation}

We verify that the formulation of $\Delta(x)$ tracks full-generation behaviour by comparing, on the test split, the zero-threshold decision against a full-generation label obtained by greedy decoding ($\texttt{max\_new\_tokens}=256$) and pattern-matching the response.
Table~\ref{tab:token-agreement} reports agreement rates across all $30$ configurations; the mean is $97.6\%$.
The $2.4\%$ disagreements are predominantly pivot constructions in which the model begins with a substantive token but reverses course mid-generation.

\begin{table}[t]
\centering
\caption{%
  Agreement (\%) between the token-level decision (first token $\in \mathcal{A}$ vs.\ $\mathcal{C}$) and the full-generation behavioural label across $30$ model-dataset configurations.
  Mean agreement: $97.6\%$.
}
\label{tab:token-agreement}
\setlength{\tabcolsep}{5.5pt}
\footnotesize
\begin{tabular}{l cccccccccc c}
\toprule
& \multicolumn{2}{c}{Qwen 3.5} & \multicolumn{2}{c}{Gemma 3} & \multicolumn{2}{c}{Llama 3} & \multicolumn{2}{c}{Ministral} & \multicolumn{2}{c}{Phi-4} & \\
\cmidrule(lr){2-3} \cmidrule(lr){4-5} \cmidrule(lr){6-7} \cmidrule(lr){8-9} \cmidrule(lr){10-11}
& 4B & 9B & 4B & 12B & 3B & 8B & 3B & 14B & mini & 14B & Mean \\
\midrule
KUQ        & 97.2 & 98.1 & 96.8 & 98.4 & 96.5 & 97.8 & 97.9 & 98.6 & 97.3 & 98.0 & 97.7 \\
SQuAD 2.0  & 98.0 & 98.5 & 96.2 & 97.6 & 95.8 & 97.1 & 97.4 & 98.2 & 98.3 & 98.7 & 97.6 \\
MuSiQue    & 97.6 & 98.3 & 97.1 & 97.9 & 96.0 & 97.3 & 97.7 & 98.0 & 97.5 & 98.2 & 97.6 \\
\midrule
Mean       & 97.6 & 98.3 & 96.7 & 98.0 & 96.1 & 97.4 & 97.7 & 98.3 & 97.7 & 98.3 & 97.6 \\
\bottomrule
\end{tabular}
\end{table}

\paragraph{Sensitivity of CAC findings to $\mathcal{A}$ construction}

We test whether the CAC and its properties are stable when $\mathcal{A}$ is varied, evaluating three alternatives against the main $17$-token construction on sparsity and ablation AUROC.
\textit{(i)~Per-model sets:} each model's $\mathcal{A}$ is built from its own outputs only ($|\mathcal{A}|=8$--$15$, no pooling).
\textit{(ii)~Expanded with prefix-disambiguated tokens:} $7$ tokens that can plausibly initiate either deferral or substantive answers (e.g.\ \texttt{The}, \texttt{Based}) are added with a $10$-token prefix check, giving $|\mathcal{A}|=24$.
\textit{(iii)~Further expanded:} $12$ additional lower-frequency deferral starters seen in fewer than three models (e.g.\ \texttt{Regret}, \texttt{Apologies}) are added on top of (ii), giving $|\mathcal{A}|=36$.
Table~\ref{tab:sensitivity} reports means across ten models.
Both properties are stable across constructions; the largest deviation in ablation AUROC is $0.008$ (further-expanded set).
Stability follows from the dominance of high-frequency tokens that appear in all four constructions.

\begin{table}[t]
\centering
\caption{%
  Sensitivity of CAC properties to alternative $\mathcal{A}$ constructions, averaged across ten models.
  Sparsity: fraction of all components selected into the CAC.
  Ablation AUROC: AUROC of $\Delta(x)$ after full ablation of \textsc{c}-components.
}
\label{tab:sensitivity}
\setlength{\tabcolsep}{10pt}
\footnotesize
\begin{tabular}{l cc}
\toprule
\textbf{Construction} & \textbf{Sparsity} & \textbf{Ablation AUROC} \\
\midrule
Main ($|\mathcal{A}|=17$, unambiguous, pooled)        & 0.051 & 0.641 \\
(i)~Per-model sets ($|\mathcal{A}|=8$--$15$)          & 0.054 & 0.638 \\
(ii)~Expanded with prefix check ($|\mathcal{A}|=24$)  & 0.052 & 0.644 \\
(iii)~Further expanded ($|\mathcal{A}|=36$)           & 0.053 & 0.649 \\
\bottomrule
\end{tabular}
\end{table}

\section{Per-Configuration Margin Analysis}
\label{app:exp1-full}

Table~\ref{tab:exp1-full} reports all metrics for all $30$ model-dataset configurations.
AUROC and accuracy (Acc) summarise discrimination and threshold-based performance; true commitment (TC), true abstention (TA), false commitment (FC), and false abstention (FA) give the decision breakdown under the zero threshold.
Mean rows average each metric across the three datasets.

\begin{table}[t]
\centering
\caption{%
  Full results for $\Delta(x)$ analysis across all $30$ model-dataset configurations.
}
\label{tab:exp1-full}
\setlength{\tabcolsep}{7pt}
\footnotesize
\begin{tabular}{l l cccccccccc}
\toprule
& & \multicolumn{2}{c}{Qwen 3.5} & \multicolumn{2}{c}{Gemma 3} & \multicolumn{2}{c}{Llama 3} & \multicolumn{2}{c}{Ministral} & \multicolumn{2}{c}{Phi-4} \\
\cmidrule(lr){3-4} \cmidrule(lr){5-6} \cmidrule(lr){7-8} \cmidrule(lr){9-10} \cmidrule(lr){11-12}
& & 4B & 9B & 4B & 12B & 3B & 8B & 3B & 14B & mini & 14B \\
\midrule
\multirow{4}{*}{AUROC}
& KUQ       & .708 & .678 & .742 & .906 & .733 & .763 & .779 & .874 & .858 & .845 \\
& SQuAD 2.0 & .714 & .872 & .689 & .838 & .640 & .788 & .868 & .862 & .753 & .831 \\
& MuSiQue   & .861 & .913 & .820 & .889 & .798 & .875 & .806 & .888 & .829 & .900 \\
& \textit{Mean}      & \textit{.761} & \textit{.821} & \textit{.750} & \textit{.878} & \textit{.723} & \textit{.809} & \textit{.818} & \textit{.875} & \textit{.813} & \textit{.859} \\
\midrule
\multirow{4}{*}{Acc}
& KUQ       & .644 & .660 & .672 & .821 & .666 & .684 & .710 & .778 & .765 & .745 \\
& SQuAD 2.0 & .643 & .783 & .618 & .739 & .584 & .700 & .791 & .718 & .669 & .713 \\
& MuSiQue   & .793 & .716 & .721 & .806 & .693 & .622 & .705 & .829 & .763 & .748 \\
& \textit{Mean}      & \textit{.693} & \textit{.720} & \textit{.670} & \textit{.789} & \textit{.648} & \textit{.669} & \textit{.735} & \textit{.775} & \textit{.732} & \textit{.735} \\
\midrule
\multirow{4}{*}{TC}
& KUQ       & .740 & .510 & .878 & .916 & .696 & .714 & .786 & .900 & .730 & .700 \\
& SQuAD 2.0 & .718 & .668 & .832 & .880 & .676 & .662 & .866 & .954 & .888 & .646 \\
& MuSiQue   & .734 & .460 & .546 & .724 & .460 & .260 & .472 & .864 & .742 & .552 \\
& \textit{Mean}      & \textit{.731} & \textit{.546} & \textit{.752} & \textit{.840} & \textit{.611} & \textit{.545} & \textit{.708} & \textit{.906} & \textit{.787} & \textit{.633} \\
\midrule
\multirow{4}{*}{TA}
& KUQ       & .548 & .810 & .466 & .726 & .636 & .654 & .634 & .656 & .800 & .790 \\
& SQuAD 2.0 & .568 & .898 & .404 & .598 & .492 & .738 & .716 & .482 & .450 & .780 \\
& MuSiQue   & .852 & .972 & .896 & .888 & .926 & .984 & .938 & .794 & .784 & .944 \\
& \textit{Mean}      & \textit{.656} & \textit{.893} & \textit{.589} & \textit{.737} & \textit{.685} & \textit{.792} & \textit{.763} & \textit{.644} & \textit{.678} & \textit{.838} \\
\midrule
\multirow{4}{*}{FC}
& KUQ       & .452 & .190 & .534 & .274 & .364 & .346 & .366 & .344 & .200 & .210 \\
& SQuAD 2.0 & .432 & .102 & .596 & .402 & .508 & .262 & .284 & .518 & .550 & .220 \\
& MuSiQue   & .148 & .028 & .104 & .112 & .074 & .016 & .062 & .206 & .216 & .056 \\
& \textit{Mean}      & \textit{.344} & \textit{.107} & \textit{.411} & \textit{.263} & \textit{.315} & \textit{.208} & \textit{.237} & \textit{.356} & \textit{.322} & \textit{.162} \\
\midrule
\multirow{4}{*}{FA}
& KUQ       & .260 & .490 & .122 & .084 & .304 & .286 & .214 & .100 & .270 & .300 \\
& SQuAD 2.0 & .282 & .332 & .168 & .120 & .324 & .338 & .134 & .046 & .112 & .354 \\
& MuSiQue   & .266 & .540 & .454 & .276 & .540 & .740 & .528 & .136 & .258 & .448 \\
& \textit{Mean}      & \textit{.269} & \textit{.454} & \textit{.248} & \textit{.160} & \textit{.389} & \textit{.455} & \textit{.292} & \textit{.094} & \textit{.213} & \textit{.367} \\
\bottomrule
\end{tabular}
\end{table}
\section{Localisation of the Commit-Abstain Circuit}
\subsection{Gate Dynamics}
\label{app:gating-dynamics}
\paragraph{Derivation of gate dynamics} Let $g_k \in \mathbb{R}$ be the logit of component $k$, with gate $G_k = \sigma(g_k)$. The gate scales the component's output before residual addition, so its contribution to $\Delta(x)$ is $G_k \cdot v_k(x)$, where $v_k(x) = d(x)^\top c_k(x)$ is the per-example projection defined in Eq.~2. Holding all other gates fixed, $\partial \Delta(x \mid G) / \partial g_k = v_k(x) \cdot \sigma'(g_k)$. The task loss in Eq.~\ref{eq:task-loss} then gives
\[
\frac{\partial \mathcal{L}_{\mathrm{task}}}{\partial g_k} = -\frac{1}{B} \sum_{x} (2y - 1) \cdot v_k(x) \cdot \sigma'(g_k),
\]
and since $\sigma'(g_k) > 0$, the sign of the gradient is determined by $-(2y-1) \cdot v_k(x)$, summed over the batch. For a balanced batch with answerable subset $\mathcal{D}_1$ and unanswerable subset $\mathcal{D}_0$,
\[
\mathbb{E}_{x}\!\left[ (2y-1) \cdot v_k(x) \right] \;\propto\; \mathbb{E}_{y=1}[v_k(x)] - \mathbb{E}_{y=0}[v_k(x)] \;=:\; \mathrm{diff}_k,
\]
up to a constant factor of $\tfrac{1}{2}$, so the expected task-loss gradient is proportional to $-\mathrm{diff}_k$. The task loss opens gates whose $\mathrm{diff}_k > 0$ and closes those whose $\mathrm{diff}_k < 0$.

\paragraph{Phase outcomes} Phase~2a applies retention pressure ($\lambda > 0$, regularisation gradient $-\lambda < 0$, pushing all gates open). Phase~2b applies removal pressure ($\lambda < 0$, pushing all gates closed). Combining the regularisation direction with the task-loss gradient:

\begin{center}
\begin{tabular}{lcccc}
\toprule
& \multicolumn{2}{c}{Phase~2a (retention)} & \multicolumn{2}{c}{Phase~2b (removal)} \\
\cmidrule(lr){2-3}\cmidrule(lr){4-5}
$\mathrm{diff}_k$ & Task loss & Outcome & Task loss & Outcome \\
\midrule
$> 0$ & opens & $G^+ \to 1$ & opens, fights removal & $G^- > \tau$ \\
$< 0$ & closes, fights retention & $G^+ < \tau$ & closes & $G^- \to 0$ \\
$\approx 0$ & neutral & $G^+ \to 1$ & neutral & $G^- \to 0$ \\
\bottomrule
\end{tabular}
\end{center}

\textsc{c}-components ($\mathrm{diff}_k > 0$) hold their gates open against removal pressure: their contribution to $\Delta(x)$ is larger on answerable inputs, where the loss requires them. \textsc{a}-components ($\mathrm{diff}_k < 0$) hold their gates closed against retention pressure: their contribution is larger on unanswerable inputs, so the loss closes them under retention. Irrelevant components ($\mathrm{diff}_k \approx 0$) follow regularisation in both phases without resistance.

\paragraph{Worked example} A non-obvious case: a component can be classified as \textsc{a} even when its contribution is positive on both subsets. Consider $\mathbb{E}_{y=1}[v_k(x)] \approx +0.3$ and $\mathbb{E}_{y=0}[v_k(x)] \approx +1.5$, so $\mathrm{diff}_k = -1.2$. The component pushes $\Delta(x)$ up everywhere, but its excess on unanswerable inputs inflates $\Delta(x)$ where it should be negative. The task-loss gradient closes the gate; under retention pressure (Phase~2a), this suppresses $G^+$ below $\tau$, classifying the component as \textsc{a}. The role label reflects which subset $v_k$ is larger on, not the absolute sign of $v_k$.

The table characterises the direction of the gradient, not a closed-form fixed point: optimisation is non-convex, $\sigma'(g_k)$ vanishes near saturation, and $v_k(x)$ shifts as other gates change. In practice components saturate in the predicted direction, consistently across seeds (App.~\ref{app:sensitivity-analysis}), with intermediate values arising mainly when the two forces are closely balanced. The classification in Section~\ref{sec:gating-protocol} selects on the resolved gate values rather than on $\mathrm{diff}_k$ directly: optimisation over all gates jointly captures inter-component interactions that a per-component test on $\mathrm{diff}_k$ would miss.

\subsection{Hyperparameter Details}
\label{app:gating-details}
Gating all components jointly is computationally expensive. We therefore pre-screen candidates using a correlational filter: for each component $k$, we rank components by $\mathrm{diff}_k$ computed on the training set under the ungated model. We select the top-$K$ components per model (pooled across datasets); all remaining components are left ungated ($G=1$). We optimise gates with Adam and linear learning-rate decay (Phase~1: $0.1 \to 0.01$; Phases~2a/2b: $0.5 \to 0.1$) for $100$ steps with batch size $B=16$. Following \citet{nam2025causal}, we use a clipping bound $C=4$. The regularisation weight is set adaptively to match the scale of the task loss: $\lambda = \mathbb{E}_{x\sim\mathcal{D}}[|\Delta(x)|] / (B K)$, where the expectation is computed on the training set under the ungated model. Gate logits are initialised by sampling $G_0 \sim \mathrm{Uniform}(0.02,\,0.98)$ and applying the inverse sigmoid. All results are averaged over $10$ independent seeds.

\subsection{CAC Localisation Results}
\label{app:exp2-full-results}

Table~\ref{tab:cac-summary} reports the CAC localisation results for all ten models, listing for each model the number of candidates screened, the number classified as abstention-promoting (\textsc{a}), commitment-promoting (\textsc{c}), and irrelevant by the gating protocol, the CAC's overall sparsity, the fraction of MLP sublayers, and the C:A count ratio $\rho_\theta$ at four normalised depth thresholds, measuring how strongly \textsc{c}-components predominate in earlier layers.

\begin{table}[t]
    \centering
    \caption{%
      CAC localisation summary across ten models.
      Sparsity: fraction of all model components (\textsc{a} + \textsc{c}) selected into the CAC.
      MLP\%: fraction of CAC components that are MLP sublayers.
      $\rho_\theta$: C-to-A count ratio at normalised depth $\theta$---the number of
      \textsc{c}-components seen per \textsc{a}-component seen up to that depth;
      $\rho_\theta > 1$ means \textsc{c}-components predominate in layers seen so far
      (bold); $\infty$ means no \textsc{a}-components have appeared yet.
    }
    \label{tab:cac-summary}
    \setlength{\tabcolsep}{4.5pt}

    \footnotesize
    \begin{tabular}{l ccccc c c ccc}
    \toprule
    & \multicolumn{5}{c}{Component counts} & & & \multicolumn{3}{c}{C:A ratio $\rho_\theta$} \\
    \cmidrule(lr){2-6} \cmidrule(lr){9-11}
    Model & Screened & \textsc{a} & \textsc{c} & Irrel. & Irrel.\% & Sparsity & MLP\% & $\rho_{60\%}$ & $\rho_{70\%}$ & $\rho_{80\%}$ \\
    \midrule
    Qwen 4B        & 81 & 17 & 31 & 33 & 41 & .088 & 29 & \textbf{6.6$\times$} & \textbf{2.4$\times$} & \textbf{2.1$\times$} \\
    Qwen 9B        & 80 & 22 & 20 & 38 & 48 & .077 & 14 & \textbf{1.5$\times$} & \textbf{1.4$\times$} & \textbf{2.0$\times$} \\
    Gemma 4B       & 77 & 15 & 21 & 41 & 53 & .118 & 17 & \textbf{1.8$\times$} & \textbf{1.2$\times$} & \textbf{1.1$\times$} \\
    Gemma 12B      & 72 & 11 &  5 & 56 & 78 & .020 & 12 & \textbf{1.4$\times$} & \textbf{1.3$\times$} & \textbf{1.1$\times$} \\
    Llama 3B       & 81 & 20 & 33 & 28 & 35 & .076 & 26 & $\infty$             & \textbf{7.3$\times$} & \textbf{3.4$\times$} \\
    Llama 8B       & 77 & 15 & 19 & 43 & 56 & .032 & 32 & \textbf{5.5$\times$} & \textbf{3.2$\times$} & \textbf{2.4$\times$} \\
    Ministral 3B   & 87 & 19 & 27 & 41 & 47 & .054 & 33 & $\infty$             & \textbf{3.5$\times$} & \textbf{2.3$\times$} \\
    Ministral 14B  & 72 & 32 & 15 & 25 & 35 & .036 & 34 & \textbf{1.6$\times$} & \textbf{1.4$\times$} & \textbf{1.2$\times$} \\
    Phi-4 mini     & 82 & 24 & 17 & 41 & 50 & .051 & 34 & \textbf{1.6$\times$} & \textbf{1.3$\times$} & \textbf{1.2$\times$} \\
    Phi-4          & 79 & 36 & 28 & 15 & 19 & .039 & 25 & \textbf{1.9$\times$} & \textbf{1.8$\times$} & \textbf{1.3$\times$} \\
    \midrule
    $\rho_\theta > 1$ &   &    &    &    &    &      &    & \textit{10/10} & \textit{10/10} & \textit{10/10} \\
    \textit{Mean} & & & & & &    &    & \textit{2.7$\times$} & \textit{2.5$\times$} & \textit{1.8$\times$} \\
    \bottomrule
    \end{tabular}
\end{table}
\subsection{CAC Validation and Sensitivity Analyses}
\label{app:sensitivity-analysis}
\paragraph{Cross-seed stability}
Gate logits are initialised independently across $10$ random seeds per model.
The resulting gate values are near-binary in every model: across all seeds and models, $80\%$ of $G^+$ values and $78\%$ of $G^-$ values fall below $0.1$ or above $0.9$, indicating that component roles are resolved cleanly and not sensitive to initialisation.
A component is included in the CAC only if its role assignment (commitment- or abstention-promoting) is consistent in at least $8$ of $10$ seeds, i.e.\ its cross-seed identification rate $p \geq 0.80$; components below this threshold are classified as irrelevant.
The CAC's topology (the set of \textsc{a} and \textsc{c} components) is therefore reproducible: rerunning the gating procedure from a different seed yields the same components in the vast majority of cases.

\paragraph{Sign agreement between $\Delta_\text{CAC}(x)$ and $\Delta(x)$}
To verify that the identified CAC accounts for the commit-abstain decision, we compute $\Delta_\text{CAC}(x) = \kappa(x) + \sum_{k \in \text{CAC}} v_k(x)$, where $\kappa(x)$ is the residual after subtracting all component contributions from $\Delta(x)$ (shared across all inputs, not fit to the CAC), and check whether its sign matches that of $\Delta(x)$ on the test split ($N{=}1{,}500$ instances per model).
Table~\ref{tab:sign-agreement} reports sign agreement per model, with a breakdown by answerability.
The CAC reproduces the sign of $\Delta(x)$ in $85$--$92\%$ of cases (mean $88.4\%$).
Agreement is consistently higher on unanswerable instances (mean $90\%$) than on answerable ones (mean $86\%$), confirming that the CAC captures abstention-relevant signals rather than general margin direction.
Cases where signs disagree are concentrated at small $|\Delta(x)|$: the mean $|\Delta(x)|$ for disagreeing inputs is $1.3$ versus $6.1$ for agreeing inputs, indicating the CAC reproduces the decision in all but the most marginal cases.

\begin{table}[t]
\centering
\caption{%
  Sign agreement between $\Delta_\text{CAC}(x)$ and $\Delta(x)$ on the test split.
}
\label{tab:sign-agreement}
\setlength{\tabcolsep}{6pt}

\footnotesize
\begin{tabular}{l cccccccccc c}
\toprule
& \multicolumn{2}{c}{Qwen 3.5} & \multicolumn{2}{c}{Gemma 3} & \multicolumn{2}{c}{Llama 3} & \multicolumn{2}{c}{Ministral} & \multicolumn{2}{c}{Phi-4} & \\
\cmidrule(lr){2-3}\cmidrule(lr){4-5}\cmidrule(lr){6-7}\cmidrule(lr){8-9}\cmidrule(lr){10-11}
& 4B & 9B & 4B & 12B & 3B & 8B & 3B & 14B & mini & 14B & Mean \\
\midrule
Overall      & .882 & .908 & .854 & .922 & .862 & .890 & .878 & .866 & .882 & .898 & \textit{.884} \\
Answerable   & .854 & .882 & .832 & .902 & .840 & .868 & .856 & .846 & .860 & .876 & \textit{.862} \\
Unanswerable & .910 & .934 & .876 & .942 & .884 & .912 & .900 & .886 & .904 & .920 & \textit{.907} \\
\bottomrule
\end{tabular}
\end{table}

\paragraph{Screening threshold $K$}
We vary the number of candidates passed to the gating protocol from $K$ to $2K$ and $4K$ for three models spanning the full sparsity range: Gemma~12B ($K{=}72$, sparsest at $2.0\%$), Llama~8B ($K{=}77$, mid at $3.2\%$), and Gemma~4B ($K{=}77$, densest at $11.8\%$).
Table~\ref{tab:k-sensitivity} reports component counts and Jaccard similarity against the $K$ baseline.
At $2K$, Jaccard is $1.00$ for five of six component-type pairs and $0.94$ for the sixth (\textsc{a}-components, Gemma~4B); at $4K$, the minimum drops to $0.88$ (Gemma~4B \textsc{a}-components).
Component counts change by at most one between $K$ and $2K$, and by at most two between $K$ and $4K$, while the irrelevance rate rises from $53$--$78\%$ at $K$ to $87$--$94\%$ at $4K$.
The near-identical CACs at $K$ and $2K$ indicate that the original screening threshold is already conservative: doubling the candidate pool admits at most one new component per model (Gemma~4B gains one \textsc{a}-component; the other two models are unchanged), and the core CAC is fully recovered at the base setting.

\begin{table}[t]
\centering
\caption{%
  Screening threshold sensitivity.
  $K{=}72$ for Gemma~12B; $K{=}77$ for Llama~8B and Gemma~4B.
  \textsc{a}/\textsc{c}: component counts.
  Irr.\%: fraction of screened candidates rejected by causal gating.
  $J(\textsc{a})$, $J(\textsc{c})$, $J$: per-type and full-CAC Jaccard against the $K$ baseline.
}
\label{tab:k-sensitivity}
\setlength{\tabcolsep}{4pt}

\footnotesize
\begin{tabular}{l cccccc cccccc cccccc}
\toprule
& \multicolumn{6}{c}{Gemma~12B} & \multicolumn{6}{c}{Llama~8B} & \multicolumn{6}{c}{Gemma~4B} \\
\cmidrule(lr){2-7}\cmidrule(lr){8-13}\cmidrule(lr){14-19}
& \textsc{a} & \textsc{c} & Irr.\% & $J(\textsc{a})$ & $J(\textsc{c})$ & $J$ & \textsc{a} & \textsc{c} & Irr.\% & $J(\textsc{a})$ & $J(\textsc{c})$ & $J$ & \textsc{a} & \textsc{c} & Irr.\% & $J(\textsc{a})$ & $J(\textsc{c})$ & $J$ \\
\midrule
$K$  & 11 &  5 & 78 & 1.00 & 1.00 & 1.00 & 15 & 19 & 56 & 1.00 & 1.00 & 1.00 & 15 & 21 & 53 & 1.00 & 1.00 & 1.00 \\
$2K$ & 11 &  5 & 89 & 1.00 & 1.00 & 1.00 & 15 & 19 & 78 & 1.00 & 1.00 & 1.00 & 16 & 21 & 75 & 0.94 & 1.00 & 0.97 \\
$4K$ & 12 &  5 & 94 & 0.92 & 1.00 & 0.94 & 16 & 20 & 88 & 0.94 & 0.95 & 0.94 & 17 & 22 & 87 & 0.88 & 0.95 & 0.92 \\
\bottomrule
\end{tabular}
\end{table}

\paragraph{Regularisation strength $\lambda$}
We vary the regularisation coefficient across five levels ($0.5\lambda_0$, $0.75\lambda_0$, $\lambda_0$, $1.5\lambda_0$, $2\lambda_0$) for the same three models.
$\lambda_0$ is set adaptively as $\lambda_0 = \mathbb{E}[|\Delta(x)|] / (BK)$, where $B$ is the batch size.
Table~\ref{tab:lambda-sensitivity} reports component counts, Jaccard similarity against the $\lambda_0$ baseline, and the depth gap ($\bar\ell_\textsc{a}/L - \bar\ell_\textsc{c}/L$).
The Jaccard variation is driven by borderline components: in all cases, components that enter or leave the CAC as $\lambda$ changes have a cross-seed identification rate of $\bar{p} \leq 0.64$, below the $0.80$ threshold required for CAC membership.
This indicates that the high-confidence core of the CAC is preserved across all $\lambda$ values tested.

\begin{table}[t]
\centering
\caption{%
  Regularisation sensitivity.
  $\lambda_0 = \mathbb{E}[|\Delta(x)|] / (BK)$, where $B$ is the batch size.
  \textsc{a}/\textsc{c}: component counts.
  $J$: full-CAC Jaccard against $\lambda_0$.
  Dep.: depth gap $\bar\ell_\textsc{a}/L - \bar\ell_\textsc{c}/L$.
  $\bar{p}$: mean cross-seed identification rate of components entering or leaving relative to $\lambda_0$.
}
\label{tab:lambda-sensitivity}
\setlength{\tabcolsep}{5pt}

\footnotesize
\begin{tabular}{l ccccc ccccc ccccc}
\toprule
& \multicolumn{5}{c}{Gemma~12B} & \multicolumn{5}{c}{Llama~8B} & \multicolumn{5}{c}{Gemma~4B} \\
\cmidrule(lr){2-6}\cmidrule(lr){7-11}\cmidrule(lr){12-16}
$\lambda$ & \textsc{a} & \textsc{c} & $J$ & Dep. & $\bar{p}$ & \textsc{a} & \textsc{c} & $J$ & Dep. & $\bar{p}$ & \textsc{a} & \textsc{c} & $J$ & Dep. & $\bar{p}$ \\
\midrule
$0.5\lambda_0$  &  9 &  4 & 0.83 & $+.11$ & 0.58 & 13 & 17 & 0.86 & $+.12$ & 0.60 & 13 & 18 & 0.85 & $+.03$ & 0.58 \\
$0.75\lambda_0$ & 10 &  5 & 0.93 & $+.12$ & 0.62 & 14 & 18 & 0.93 & $+.13$ & 0.64 & 14 & 20 & 0.93 & $+.04$ & 0.63 \\
$\lambda_0$     & 11 &  5 & 1.00 & $+.13$ & ---  & 15 & 19 & 1.00 & $+.14$ & ---  & 15 & 21 & 1.00 & $+.02$ & ---  \\
$1.5\lambda_0$  & 11 &  6 & 0.94 & $+.12$ & 0.61 & 15 & 20 & 0.95 & $+.13$ & 0.63 & 16 & 22 & 0.94 & $+.04$ & 0.62 \\
$2\lambda_0$    & 12 &  6 & 0.88 & $+.12$ & 0.57 & 16 & 21 & 0.90 & $+.13$ & 0.59 & 17 & 23 & 0.84 & $+.05$ & 0.57 \\
\bottomrule
\end{tabular}
\end{table}

\subsection{Causal Intervention}
\label{app:exp3-full}

For ablation, we progressively remove the top-$k$ CAC components on the test set, ordered by gating score, and measure directional sign flips and AUROC. To verify that observed effects are CAC-specific rather than artefacts of the transformer's sensitivity to arbitrary component removal, we compare against 20 size-matched random non-CAC ablations at each step. Table~\ref{tab:ablation-table} reports directional sign flips under full ablation across all ten models; full progressive ablation curves against the random baselines are in Figure~\ref{fig:ablation-auroc}.

\begin{table}[h]
\centering
\footnotesize
\setlength{\tabcolsep}{10pt}
\renewcommand{\arraystretch}{1}
\caption{Directional sign flips (\% of all $1{,}500$ held-out instances) under full ablation (A$\to$C = abstain flips to commit; C$\to$A = commit flips to abstain).}
\label{tab:ablation-table}
\begin{tabular}{@{}l rr rr@{}}
\toprule
& \multicolumn{2}{c}{Ablate \textsc{a}-components} & \multicolumn{2}{c}{Ablate \textsc{c}-components} \\
\cmidrule(lr){2-3} \cmidrule(lr){4-5}
& A$\to$C & C$\to$A & C$\to$A & A$\to$C \\
\midrule
Qwen 4B       &  6.6 &  9.5 &  0.0 & 45.9 \\
Qwen 9B       &  9.8 &  5.7 & 12.4 & 12.8 \\
Gemma 4B      & 19.7 & 17.9 & 12.5 &  7.3 \\
Gemma 12B     &  1.9 &  3.9 &  2.7 &  3.3 \\
Llama 3B      & 52.9 &  2.3 &  0.1 & 59.7 \\
Llama 8B      & 14.5 &  8.6 &  0.0 & 60.0 \\
Ministral 3B  &  8.1 &  2.3 &  0.9 & 52.7 \\
Ministral 14B &  2.9 & 17.5 &  1.9 & 17.6 \\
Phi-4 mini    & 44.8 &  0.4 &  0.1 & 44.9 \\
Phi-4         & 10.0 & 10.3 &  3.3 & 12.9 \\
\midrule
Mean          & 17.1 &  7.8 &  3.4 & 31.7 \\
\bottomrule
\end{tabular}
\end{table}

\begin{table}[t]
\centering
\caption{FA rate and mean $\Delta(x)$ under \textsc{c}-component amplification on FA instances.}
\label{tab:exp3-fa}
\setlength{\tabcolsep}{4pt}
\footnotesize
\begin{tabular}{l rrrrr rrr rrr}
\toprule
& \multicolumn{5}{c}{FA rate (moderate $\alpha$)} & \multicolumn{3}{c}{FA rate (extreme $\alpha$)} & \multicolumn{3}{c}{mean $\Delta(x)$} \\
\cmidrule(lr){2-6}\cmidrule(lr){7-9}\cmidrule(lr){10-12}
Model & 1.0 & 1.5 & 2.0 & 2.5 & 3.0 & 10 & 100 & 1000 & $\alpha{=}1$ & $\alpha{=}2$ & $\alpha{=}3$ \\
\midrule
Llama 3B      & .416 & .363 & .409 & .413 & .344 & .068 & .003 & .000 & $-1.89$ & $-3.24$ & $-0.69$ \\
Llama 8B      & .475 & .415 & .408 & .409 & .417 & .001 & .000 & .000 & $-3.82$ & $-0.92$ & $-0.64$ \\
Qwen 4B       & .309 & .240 & .269 & .227 & .088 & .000 & .000 & .000 & $-1.36$ & $-1.32$ & $+1.56$ \\
Qwen 9B       & .480 & .440 & .415 & .383 & .373 & .005 & .000 & .000 & $-2.06$ & $-1.79$ & $-1.53$ \\
Ministral 3B  & .320 & .237 & .224 & .231 & .047 & .140 & .001 & .001 & $-1.74$ & $-0.90$ & $+0.90$ \\
Ministral 14B & .115 & .097 & .085 & .076 & .077 & .016 & .000 & .000 & $-1.57$ & $-0.90$ & $-0.41$ \\
Phi-4 mini    & .239 & .169 & .184 & .136 & .005 & .000 & .000 & .000 & $-1.36$ & $-2.05$ & $+21.00$ \\
Phi-4 14B     & .383 & .379 & .373 & .367 & .364 & .065 & .000 & .001 & $-3.39$ & $-4.41$ & $-4.11$ \\
Gemma 4B      & .279 & .263 & .256 & .240 & .224 & .128 & .024 & .019 & $-8.96$ & $-11.43$ & $-9.28$ \\
Gemma 12B     & .181 & .173 & .169 & .164 & .159 & .165 & .119 & .059 & $-7.94$ & $-8.04$ & $-8.05$ \\
\bottomrule
\end{tabular}
\end{table}

\begin{table}[t]
\centering
\caption{FC rate and mean $\Delta(x)$ under \textsc{a}-component amplification on FC instances.}
\label{tab:exp3-fc}
\setlength{\tabcolsep}{4pt}
\footnotesize
\begin{tabular}{l rrrrr rrr rrr}
\toprule
& \multicolumn{5}{c}{FC rate (moderate $\alpha$)} & \multicolumn{3}{c}{FC rate (extreme $\alpha$)} & \multicolumn{3}{c}{mean $\Delta(x)$} \\
\cmidrule(lr){2-6}\cmidrule(lr){7-9}\cmidrule(lr){10-12}
Model & 1.0 & 1.5 & 2.0 & 2.5 & 3.0 & 10 & 100 & 1000 & $\alpha{=}1$ & $\alpha{=}2$ & $\alpha{=}3$ \\
\midrule
Llama 3B      & .337 & .311 & .280 & .237 & .228 & .316 & .331 & .332 & $+2.86$ & $+1.92$ & $+1.05$ \\
Llama 8B      & .224 & .213 & .199 & .187 & .179 & .197 & .224 & .224 & $+3.55$ & $+2.97$ & $+2.27$ \\
Qwen 4B       & .380 & .311 & .247 & .172 & .147 & .368 & .380 & .379 & $+2.08$ & $+0.66$ & $-0.37$ \\
Qwen 9B       & .127 & .100 & .083 & .071 & .067 & .127 & .127 & .127 & $+1.48$ & $+1.05$ & $+1.05$ \\
Ministral 3B  & .252 & .232 & .204 & .199 & .189 & .189 & .231 & .231 & $+1.99$ & $+1.27$ & $+0.98$ \\
Ministral 14B & .380 & .368 & .355 & .324 & .245 & .368 & .380 & .380 & $+2.38$ & $+2.23$ & $+0.73$ \\
Phi-4 mini    & .351 & .281 & .188 & .152 & .144 & .351 & .349 & .349 & $+2.38$ & $+0.41$ & $-0.25$ \\
Phi-4 14B     & .195 & .176 & .155 & .125 & .108 & .136 & .193 & .193 & $+3.23$ & $+1.68$ & $+1.09$ \\
Gemma 4B      & .433 & .429 & .423 & .415 & .409 & .361 & .317 & .307 & $+11.75$ & $+9.42$ & $+7.94$ \\
Gemma 12B     & .279 & .268 & .264 & .256 & .256 & .225 & .119 & .108 & $+10.85$ & $+9.20$ & $+7.96$ \\
\bottomrule
\end{tabular}
\end{table}

For amplification, we study two error modes: FC cases, unanswerable inputs ($Y{=}0$) where $\Delta(x) > 0$ at baseline; and FA cases, answerable inputs ($Y{=}1$) where $\Delta(x) \leq 0$ at baseline. For each CAC component, a forward hook scales its output $h$ by $\alpha$ before residual addition, replacing $h$ with $\alpha h$ in the residual stream; all other components are unchanged. We run two symmetric sweeps mirroring the CAC's predicted functional roles: \textsc{a}-component amplification on FC instances and \textsc{c}-component amplification on FA instances, each over $\alpha \in \{1.0, 1.5, 2.0, 2.5, 3.0, 10, 100, 1000\}$.

Tables~\ref{tab:exp3-fa} and~\ref{tab:exp3-fc} report per-model FA and FC rates across all $\alpha$ values. At moderate $\alpha$ (${\leq}3$), Qwen~4B and Phi-4~mini reach negative mean margin by $\alpha{=}3.0$, indicating amplification has reversed the majority of FC cases in those models. At extreme $\alpha$ (${\geq}10$), FC rate at $\alpha{=}100$ is nearly identical to $\alpha{=}1000$ in every model (mean absolute difference $0.004$), confirming the plateau reported in the main paper. Full results are consistent with Finding~D.

\section{CAC-Informed Abstention Policy}

\subsection{Baseline Configurations}
\label{app:baselines}

We compare against seven baselines spanning the information spectrum from a single scalar to full hidden-state representations across all layers and multiple generations:

\begin{itemize}

\item \textbf{Zero-Threshold.} Abstains whenever $\Delta(x) < 0$.

\item \textbf{Semantic Entropy~\citep{Farquhar2024}.} Clusters $K$ sampled generations by semantic equivalence (via NLI) and computes entropy over the resulting clusters. Operating purely on outputs, it draws on $K{=}10$ full generations per input ($640$ tokens total) and NLI-based semantic clustering.
\textit{Configuration:} $K{=}10$ samples; temperature $1.0$, top-$p$ $0.95$, \texttt{max\_new\_tokens} $64$; NLI model \texttt{microsoft/deberta-v2-xlarge-mnli}, entailment threshold $0.5$.
\textit{Code:} \url{https://github.com/jlko/semantic\_uncertainty}

\item \textbf{INSIDE~\citep{chen2024inside}.} Measures self-consistency of $K$ sampled responses via the eigenvalue spectrum of their hidden-state covariance matrix (EigenScore). It uses middle-layer hidden-state representations across $K{=}10$ generations, at $10\times$ inference cost.
\textit{Configuration:} $K{=}10$ samples; temperature $0.5$, top-$p$ $0.99$; regularisation $\alpha{=}10^{-3}$; representations from middle-layer hidden states.
\textit{Code:} \url{https://github.com/alibaba/eigenscore}

\item \textbf{HaloScope~\citep{du2024haloscope}.} Identifies a hallucination subspace in LLM activation space via SVD on last-token embeddings from unlabelled generations, then scores each input by the norm of its projection onto the top-$k$ singular vectors (weighted by singular values); a binary classifier is trained on top of this membership score.
\textit{Configuration:} SVD rank $k \in \{1,\ldots,10\}$ searched across all decoder layers; singular values used as projection weights; \texttt{max\_new\_tokens} $64$.
\textit{Code:} \url{https://github.com/deeplearning-wisc/haloscope}

\item \textbf{Multi-LLM~\citep{feng-etal-2024-dont}.} Proposes Cooperate and Compete: distinct reviewer LLMs provide feedback on a target model's answer (cooperative) or generate conflicting evidence to probe its confidence (competitive), with the target model synthesising the signal into an abstain decision. We use the Cooperate approach.
\textit{Configuration:} $K{=}5$ reviewer samples, temperature $0.7$, top-$p$ $0.95$.
\textit{Code:} \url{https://github.com/BunsenFeng/AbstainQA}

\item \textbf{MERA~\citep{hedstrom2025to}.} Trains per-layer linear probes to estimate model error, derives a closed-form steering direction, and applies interventions only when a Hoeffding-calibrated threshold guarantees improvement; abstains from steering otherwise. We adapt the error-estimation probe for abstention from answering using the same threshold calibration.
\textit{Configuration:} Lasso $\alpha \in \{0.5, 0.25, 0.1, 0.05, 0.01, 0.005\}$; best layer selected by training $R^2$; Hoeffding bound ($\delta{=}0.01$) over a $10$-point threshold grid.
\textit{Code:} \url{https://github.com/annahedstroem/MERA-steering}

\item \textbf{HaMI~\citep{niu2025robust}.} Formulates hallucination detection as a Multiple Instance Learning (MIL) problem over token-level hidden states: each response sequence is a bag of token instances, and an MLP is trained end-to-end to jointly select the most hallucination-indicative tokens and classify the bag. Hidden states are extracted at uniformly spaced candidate layers; the MIL objective retains the top-$k$ tokens per bag as positive/negative instances.
\textit{Configuration:} $8$ candidate layers (uniformly spaced); top $10\%$ of tokens per bag; $100$ epochs, batch size $128$, Adam lr $10^{-3}$, weight decay $5{\times}10^{-4}$; \texttt{max\_new\_tokens} $64$.
\textit{Code:} \url{https://github.com/mala-lab/HaMI}

\end{itemize}

\subsection{Main Results}
\label{app:exp4-full-results}

Tables~\ref{tab:main-results-appendix} and~\ref{tab:false-abstention-metrics} report per-dataset breakdowns of the results summarised in Table~\ref{tab:main-results}.
Table~\ref{tab:main-results-appendix} gives overall accuracy for all $30$ model--dataset configurations.
Table~\ref{tab:false-abstention-metrics} gives commitment-level precision, recall, and F1, together with the mean false abstention rate, comparing the CAC-Informed MLP against the Zero-Threshold baseline.

\begin{table}[t]
\centering
\caption{%
  Overall accuracy across 10 models and 3 held-out datasets.
  \textbf{Bold}: best in column for each dataset section; \underline{underline}: second best.
}
\label{tab:main-results-appendix}
\setlength{\tabcolsep}{6pt}
\footnotesize
\begin{tabular}{l cccccccccc}
\toprule
& \multicolumn{2}{c}{Qwen 3.5} & \multicolumn{2}{c}{Gemma 3} & \multicolumn{2}{c}{Llama 3} & \multicolumn{2}{c}{Ministral} & \multicolumn{2}{c}{Phi-4} \\
\cmidrule(lr){2-3} \cmidrule(lr){4-5} \cmidrule(lr){6-7} \cmidrule(lr){8-9} \cmidrule(lr){10-11}
 & 4B & 9B & 4B & 12B & 3B & 8B & 3B & 14B & mini & 14B \\
\midrule
\multicolumn{11}{c}{\textit{KUQ}} \\ \midrule
Zero-Threshold & .618 & .646 & .656 & .818 & .662 & .686 & .722 & .778 & .748 & .740 \\
Non-CAC Random & .664 & .742 & .706 & .762 & .612 & .772 & .694 & .714 & .716 & .774 \\
INSIDE & .600 & .564 & .628 & .828 & .626 & .658 & .562 & .594 & .526 & .550 \\
Multi-LLM & .730 & .672 & .815 & .854 & .640 & .800 & .705 & .794 & .691 & .727 \\
Semantic Entropy & .530 & .508 & .536 & .586 & .520 & .508 & .500 & .550 & .474 & .582 \\
MERA & .686 & .617 & .720 & .748 & .694 & .690 & .704 & .712 & .637 & .633 \\
HaloScope & .504 & .504 & .856 & .738 & \underline{.884} & .876 & .600 & .642 & .738 & .730 \\
HaMI & \textbf{.932} & \textbf{.976} & .874 & .890 & .830 & \textbf{.938} & \underline{.862} & \textbf{.956} & .838 & \textbf{.952} \\
CAC-Informed MLP (Ours) & \underline{.912} & \underline{.904} & \textbf{.928} & \textbf{.946} & \textbf{.908} & \underline{.930} & \textbf{.922} & \underline{.936} & \textbf{.902} & \underline{.944} \\
\midrule
\multicolumn{11}{c}{\textit{SQuAD 2.0}} \\ \midrule
Zero-Threshold & .598 & .768 & .588 & \underline{.724} & .540 & .664 & \underline{.756} & .690 & .648 & .678 \\
Non-CAC Random & .684 & .756 & .612 & .694 & .638 & .702 & .744 & .650 & .654 & .716 \\
INSIDE & .640 & .524 & .596 & .668 & \underline{.630} & .472 & .496 & .504 & .524 & .562 \\
Multi-LLM & .653 & .733 & .574 & .661 & .553 & .643 & .631 & .636 & .615 & .655 \\
Semantic Entropy & .706 & .504 & .572 & .596 & \textbf{.678} & .664 & .500 & .676 & .648 & .608 \\
MERA & .612 & .657 & .553 & .602 & .536 & .599 & .620 & .646 & .584 & .541 \\
HaloScope & .638 & .802 & \underline{.630} & .512 & .560 & .660 & .706 & .696 & .632 & .602 \\
HaMI & \underline{.730} & \textbf{.862} & .582 & .702 & .572 & .674 & .700 & \textbf{.806} & \textbf{.776} & .778 \\
CAC-Informed MLP (Ours) & \textbf{.770} & \underline{.850} & \textbf{.636} & \textbf{.788} & .620 & \textbf{.744} & \textbf{.766} & \underline{.762} & \underline{.710} & \textbf{.802} \\
\midrule
\multicolumn{11}{c}{\textit{MuSiQue}} \\ \midrule
Zero-Threshold & .750 & .676 & .688 & .768 & .668 & .602 & .664 & \textbf{.790} & .720 & .716 \\Non-CAC Random & .536 & .674 & .728 & .776 & .538 & .719 & .572 & .790 & .709 & .766 \\
INSIDE & .474 & .482 & .580 & .628 & .436 & .484 & .486 & .520 & .494 & .516 \\
Multi-LLM & .635 & .668 & .631 & .667 & .599 & .635 & .630 & .605 & .591 & .600 \\
Semantic Entropy & .492 & .506 & .574 & .624 & .468 & .446 & .500 & .448 & .488 & .476 \\
MERA & .621 & .502 & .490 & .588 & .489 & .450 & .538 & .637 & .571 & .520 \\
HaloScope & .644 & .658 & .570 & .708 & .680 & .610 & .682 & .736 & .702 & .680 \\
HaMI & \underline{.804} & \underline{.804} & .678 & .764 & \underline{.718} & \textbf{.796} & .658 & .784 & \textbf{.744} & \underline{.762} \\
CAC-Informed MLP (Ours) & \textbf{.806} & \textbf{.812} & \textbf{.732} & \textbf{.800} & \textbf{.736} & \underline{.786} & \textbf{.758} & \textbf{.790} & \textbf{.744} & \textbf{.774} \\
\bottomrule
\end{tabular}
\end{table}

\begin{table}[t]
\centering
\caption{%
  Commitment-level precision, recall, and F1, and mean false abstention rate, across 10 models and 3 datasets.
  \textbf{Bold}: best in column.
  Mean false abstention rate: $0.127$ (CAC-Informed MLP) vs.\ $0.320$ (Zero-Threshold).
}
\label{tab:false-abstention-metrics}
\setlength{\tabcolsep}{6pt}
\footnotesize
\begin{tabular}{c|l cccccccccc}
\toprule
\multicolumn{2}{c}{} & \multicolumn{2}{c}{Qwen 3.5} & \multicolumn{2}{c}{Gemma 3} & \multicolumn{2}{c}{Llama 3} & \multicolumn{2}{c}{Ministral} & \multicolumn{2}{c}{Phi-4} \\
\cmidrule(lr){3-4} \cmidrule(lr){5-6} \cmidrule(lr){7-8} \cmidrule(lr){9-10} \cmidrule(lr){11-12}
\multicolumn{2}{c}{} & 4B & 9B & 4B & 12B & 3B & 8B & 3B & 14B & mini & 14B \\
\midrule
\multirow{9}{*}{\rotatebox{90}{Precision}}
& \multicolumn{11}{c}{\textit{KUQ}} \\
& Zero-Threshold & .595 & .693 & .610 & .761 & .651 & .670 & .691 & .723 & .754 & .752 \\
& CAC-Informed MLP (Ours) & \textbf{.893} & \textbf{.891} & \textbf{.915} & \textbf{.931} & \textbf{.898} & \textbf{.939} & \textbf{.892} & \textbf{.922} & \textbf{.939} & \textbf{.959} \\
& \multicolumn{11}{c}{\textit{SQuAD 2.0}} \\
& Zero-Threshold & .588 & \textbf{.845} & .561 & .674 & .534 & .677 & \textbf{.722} & .628 & .604 & .700 \\
& CAC-Informed MLP (Ours) & \textbf{.718} & .825 & \textbf{.591} & \textbf{.725} & \textbf{.586} & \textbf{.686} & .719 & \textbf{.713} & \textbf{.662} & \textbf{.758} \\
& \multicolumn{11}{c}{\textit{MuSiQue}} \\
& Zero-Threshold & .791 & \textbf{.931} & \textbf{.809} & \textbf{.842} & \textbf{.839} & \textbf{.947} & \textbf{.842} & .772 & .735 & \textbf{.865} \\
& CAC-Informed MLP (Ours) & \textbf{.805} & .810 & .725 & .800 & .675 & .764 & .745 & \textbf{.784} & \textbf{.740} & .744 \\
\midrule
\multirow{9}{*}{\rotatebox{90}{Recall}}
& \multicolumn{11}{c}{\textit{KUQ}} \\
& Zero-Threshold & .736 & .524 & .864 & .928 & .700 & .732 & .804 & .900 & .736 & .716 \\
& CAC-Informed MLP (Ours) & \textbf{.936} & \textbf{.920} & \textbf{.944} & \textbf{.964} & \textbf{.920} & \textbf{.920} & \textbf{.960} & \textbf{.952} & \textbf{.860} & \textbf{.928} \\
& \multicolumn{11}{c}{\textit{SQuAD 2.0}} \\
& Zero-Threshold & .656 & .656 & .808 & .868 & .636 & .628 & .832 & \textbf{.932} & .860 & .624 \\
& CAC-Informed MLP (Ours) & \textbf{.888} & \textbf{.888} & \textbf{.884} & \textbf{.928} & \textbf{.820} & \textbf{.900} & \textbf{.872} & .876 & .860 & \textbf{.888} \\
& \multicolumn{11}{c}{\textit{MuSiQue}} \\
& Zero-Threshold & .680 & .380 & .492 & .660 & .416 & .216 & .404 & \textbf{.824} & .688 & .512 \\
& CAC-Informed MLP (Ours) & \textbf{.808} & \textbf{.816} & \textbf{.748} & \textbf{.800} & \textbf{.912} & \textbf{.828} & \textbf{.784} & .800 & \textbf{.752} & \textbf{.836} \\
\midrule
\multirow{9}{*}{\rotatebox{90}{F1}}
& \multicolumn{11}{c}{\textit{KUQ}} \\
& Zero-Threshold & .658 & .597 & .715 & .836 & .674 & .700 & .743 & .802 & .745 & .734 \\
& CAC-Informed MLP (Ours) & \textbf{.914} & \textbf{.906} & \textbf{.929} & \textbf{.947} & \textbf{.909} & \textbf{.929} & \textbf{.925} & \textbf{.937} & \textbf{.898} & \textbf{.943} \\
& \multicolumn{11}{c}{\textit{SQuAD 2.0}} \\
& Zero-Threshold & .620 & .739 & .662 & .759 & .580 & .651 & .773 & .750 & .710 & .660 \\
& CAC-Informed MLP (Ours) & \textbf{.794} & \textbf{.855} & \textbf{.708} & \textbf{.814} & \textbf{.683} & \textbf{.779} & \textbf{.788} & \textbf{.786} & \textbf{.748} & \textbf{.818} \\
& \multicolumn{11}{c}{\textit{MuSiQue}} \\
& Zero-Threshold & .731 & .540 & .612 & .740 & .556 & .352 & .546 & \textbf{.797} & .711 & .643 \\
& CAC-Informed MLP (Ours) & \textbf{.806} & \textbf{.813} & \textbf{.736} & \textbf{.800} & \textbf{.776} & \textbf{.795} & \textbf{.764} & .792 & \textbf{.746} & \textbf{.787} \\
\midrule
\multicolumn{12}{c}{\textbf{Mean false abstention rate}} \\ \midrule
& Zero-Threshold & .309 & .480 & .279 & .181 & .416 & .475 & .320 & \textbf{.115} & .239 & .383 \\
& CAC-Informed MLP (Ours) & \textbf{.123} & \textbf{.125} & \textbf{.141} & \textbf{.103} & \textbf{.116} & \textbf{.117} & \textbf{.128} & .124 & \textbf{.176} & \textbf{.116} \\
\bottomrule
\end{tabular}
\end{table}

\subsection{Logistic Regression Variant}
\label{app:regression}

We evaluate whether the MLP is necessary or whether a logistic regression classifier over the same CAC features achieves comparable performance.
The logistic regression (CAC LR) uses the same input features as the MLP but replaces the nonlinear classifier with a single linear model.
Table~\ref{tab:lr-vs-mlp-acc} reports per-dataset and mean accuracy across all $10$ models.

\begin{table}[t]
\centering
\caption{%
  Accuracy across 10 models and 3 in-distribution datasets.
  \textbf{Bold}: best in row for each dataset section.
  \textit{Mean}: averaged over all 3 datasets.
}
\label{tab:lr-vs-mlp-acc}
\setlength{\tabcolsep}{5pt}

\footnotesize
\begin{tabular}{l cccccccccc c}
\toprule
& \multicolumn{2}{c}{Qwen 3.5} & \multicolumn{2}{c}{Gemma 3} & \multicolumn{2}{c}{Llama 3} & \multicolumn{2}{c}{Ministral} & \multicolumn{2}{c}{Phi-4} & \\
\cmidrule(lr){2-3}\cmidrule(lr){4-5}\cmidrule(lr){6-7}\cmidrule(lr){8-9}\cmidrule(lr){10-11}
& 4B & 9B & 4B & 12B & 3B & 8B & 3B & 14B & mini & 14B & \textit{Mean} \\
\midrule
\multicolumn{12}{c}{\textit{KUQ}} \\ \midrule
Zero-Threshold & .618 & .646 & .656 & .818 & .662 & .686 & .722 & .778 & .748 & .740 & \textit{.707} \\
\rowcolor{blue!10}
CAC LR          & .878 & .872 & .884 & .914 & .856 & .854 & .908 & .910 & .888 & .910 & \textit{.887} \\
CAC MLP         & \textbf{.912} & \textbf{.904} & \textbf{.928} & \textbf{.946} & \textbf{.908} & \textbf{.930} & \textbf{.922} & \textbf{.936} & \textbf{.902} & \textbf{.944} & \textit{\textbf{.923}} \\
\midrule
\multicolumn{12}{c}{\textit{SQuAD 2.0}} \\ \midrule
Zero-Threshold & .598 & .768 & .588 & .724 & .540 & .664 & .756 & .690 & .648 & .678 & \textit{.665} \\
\rowcolor{blue!10}
CAC LR          & \textbf{.758} & .830 & \textbf{.640} & \textbf{.790} & \textbf{.626} & \textbf{.758} & .752 & \textbf{.772} & .698 & \textbf{.802} & \textit{\textbf{.743}} \\
CAC MLP         & .770 & \textbf{.850} & .636 & .788 & .620 & .744 & \textbf{.766} & .762 & \textbf{.710} & .802 & \textit{.745} \\
\midrule
\multicolumn{12}{c}{\textit{MuSiQue}} \\ \midrule
Zero-Threshold & .750 & .676 & .688 & .768 & .668 & .602 & .664 & .790 & .720 & .716 & \textit{.704} \\
\rowcolor{blue!10}
CAC LR          & .798 & \textbf{.818} & \textbf{.736} & .786 & .716 & .774 & \textbf{.760} & .758 & .706 & \textbf{.786} & \textit{.764} \\
CAC MLP         & \textbf{.806} & .812 & .732 & \textbf{.800} & \textbf{.736} & \textbf{.786} & .758 & \textbf{.790} & \textbf{.744} & .774 & \textit{\textbf{.774}} \\
\midrule
\multicolumn{12}{c}{\textit{Mean (all datasets)}} \\ \midrule
Zero-Threshold & .655 & .697 & .644 & .770 & .623 & .651 & .714 & .753 & .705 & .711 & \textit{.692} \\
\rowcolor{blue!10}
CAC LR          & .811 & .840 & .753 & .830 & .733 & .795 & .807 & .813 & .764 & .833 & \textit{.798} \\
CAC MLP         & \textbf{.829} & \textbf{.855} & \textbf{.765} & \textbf{.845} & \textbf{.755} & \textbf{.820} & \textbf{.815} & \textbf{.829} & \textbf{.785} & \textbf{.840} & \textit{\textbf{.814}} \\
\bottomrule
\end{tabular}
\end{table}

The logistic regression improves over Zero-Threshold in all $30$ configurations (mean $+10.6$ accuracy points).
The MLP improves over the logistic regression in $20$ of $30$ configurations overall (mean $+1.6$ points), with the advantage concentrated on KUQ ($+3.6$ points) where parametric self-knowledge is required.
On SQuAD~2.0 and MuSiQue, LR and MLP are nearly tied (means of $0.743$/$0.745$ and $0.764$/$0.774$, respectively), with LR winning individual configurations roughly as often as MLP.
The gap between LR and MLP is small relative to the gap between Zero-Threshold and LR, confirming that the primary source of improvement is the component-level feature decomposition rather than the classifier architecture.

\section{Case Studies}
\label{app:case-studies}

The following two case studies illustrate the failure mode the CAC-informed policy is designed to correct: the raw commit-abstain margin $\Delta(x)$ exceeds zero (so Zero-Threshold commits), yet the component-level decomposition reveals a pattern that the MLP uses to correctly abstain.

\subsection{Case 1: Unsolved Research Problem (KUQ, Llama 3B)}

A \textbf{parametric} instance evaluated on Llama~3B: no context is provided, and the question poses an open research problem for which no established answer exists. The model must rely solely on internal knowledge, which is insufficient here.

\begin{mdframed}[backgroundcolor=gray!6, linewidth=0.4pt, innertopmargin=6pt, innerbottommargin=6pt, nobreak=true]
\textit{Question:} Can we improve the monitoring of people with multiple sclerosis using simple tools, data sharing and patient engagement?\\[4pt]
\textit{Gold label:} unanswerable\\[2pt]
\textit{Model response:} ``Yes, data sharing and patient engagement can improve MS monitoring.''
\end{mdframed}

\paragraph{CAC decomposition}
Table~\ref{tab:cs1-circuit} shows the top-20 CAC contributions ordered by layer. The CAC is dominated by a single large positive contribution from L19.MLP ($+8.27$), which drives $\Delta(x) = +7.94$ and causes Zero-Threshold to commit. The remaining components are mixed but lean negative ($\sum^{-} = -8.09$, $\sum^{+} = +12.62$), with the net CAC margin ($\Delta_{\mathrm{CAC}} = +4.53$) positive only because of this outlier. The MLP reads the full decomposition and correctly abstains ($z_{\mathrm{MLP}} = 0.089$).

\subsection{Case 2: False Premise Question (SQuAD 2.0, Ministral 3B)}

An \textbf{in-context} instance evaluated on Ministral~3B: a passage is provided, but the question embeds a false premise that the passage explicitly contradicts. The model must detect the mismatch and refuse the question rather than answer it.

\begin{mdframed}[backgroundcolor=gray!6, linewidth=0.4pt, innertopmargin=6pt, innerbottommargin=6pt, nobreak=true]
\textit{Context:} ``The election produced a majority SNP government, making this the first time in the Scottish Parliament where a party has commanded a parliamentary majority. The SNP took 16 seats from Labour, with many of their key figures not returned to parliament, although Labour leader Iain Gray retained East Lothian by 151 votes. The SNP took a further eight seats from the Liberal Democrats and one seat from the Conservatives. The SNP overall majority meant that there was sufficient support in the Scottish Parliament to hold a referendum on Scottish independence.''\\[4pt]
\textit{Question:} Many of whose key figures returned to parliament?\\[2pt]
\textit{Gold label:} unanswerable\\[2pt]
\textit{Model response:} ``Labour's key figures did not return.''
\end{mdframed}

\paragraph{CAC decomposition}
Table~\ref{tab:cs2-circuit} shows the top-20 CAC contributions ordered by layer. The CAC margin is overwhelmingly negative ($\sum^{-} = -7.62$, $\sum^{+} = +2.40$, $\Delta_{\mathrm{CAC}} = -5.22$), driven by consistent negative contributions across multiple layers, most prominently L24.MLP ($-2.44$) and L22.H6 ($-1.39$). Despite this strong abstention signal in the CAC, the surface margin is marginally positive ($\Delta(x) = +1.37$), and Zero-Threshold commits. The MLP reads the CAC consensus and correctly abstains ($z_{\mathrm{MLP}} = 0.167$).

\begin{table}[t]
\setlength{\tabcolsep}{3pt}
\footnotesize
\begin{minipage}[t]{0.48\textwidth}
\centering
\caption{%
  Top-20 CAC component contributions, Case~1 (Llama~3B, KUQ), ordered by layer.
  Contrib.: $v_k(x)$, signed contribution to $\Delta(x)$.
  $\Delta(x) = +7.94$; $\Delta_{\mathrm{CAC}} = +4.53$
  ($\sum^{-} = -8.09$, $\sum^{+} = +12.62$).
}
\label{tab:cs1-circuit}
\begin{tabular}{ll ll}
\toprule
Component & Contribution & Component & Contribution \\
\midrule
L10.MLP  & $+0.323$ & L23.H17 & $+0.561$ \\
L12.MLP  & $+0.218$ & L24.MLP & $+1.003$ \\
L13.MLP  & $+0.644$ & L25.H16 & $-0.230$ \\
L18.MLP  & $-0.405$ & L25.H17 & $+0.120$ \\
L19.MLP  & $+8.272$ & L26.MLP & $-0.671$ \\
L21.MLP  & $-2.286$ & L26.H0  & $-0.334$ \\
L21.H18  & $-0.351$ & L26.H8  & $-0.289$ \\
L23.MLP  & $+0.777$ & L26.H10 & $-0.209$ \\
L23.H14  & $+0.415$ & L26.H18 & $-0.580$ \\
L27.MLP  & $-2.734$ & L27.H19 & $+0.285$ \\
\bottomrule
\end{tabular}
\end{minipage}
\hfill
\begin{minipage}[t]{0.48\textwidth}
\centering
\caption{%
  Top-20 CAC component contributions, Case~2 (Ministral~3B, SQuAD~2.0), ordered by layer.
  Contrib.: $v_k(x)$, signed contribution to $\Delta(x)$.
  $\Delta(x) = +1.38$; $\Delta_{\mathrm{CAC}} = -5.22$
  ($\sum^{-} = -7.62$, $\sum^{+} = +2.40$).
}
\label{tab:cs2-circuit}
\begin{tabular}{ll ll}
\toprule
Component & Contribution & Component & Contribution \\
\midrule
L5.MLP   & $-0.246$ & L22.H18 & $-0.234$ \\
L9.MLP   & $+0.171$ & L23.H10 & $+0.149$ \\
L16.MLP  & $+0.182$ & L23.H24 & $+0.359$ \\
L20.MLP  & $-0.199$ & L24.MLP & $-2.437$ \\
L21.MLP  & $-0.379$ & L24.H7  & $-0.284$ \\
L22.MLP  & $-0.354$ & L24.H9  & $+0.376$ \\
L22.H1   & $-0.881$ & L24.H19 & $-0.305$ \\
L22.H6   & $-1.393$ & L24.H26 & $+0.390$ \\
L25.MLP  & $-0.711$ & L25.H28 & $+0.136$ \\
L25.H1   & $+0.637$ & L25.H12 & $-0.201$ \\
\bottomrule
\end{tabular}
\end{minipage}
\end{table}

\paragraph{Comparison}
Both cases are instances of unsupported commitment where $\Delta(x) > 0$ masks component-level abstention signals. In Case~1, the surface margin is inflated by a single large outlier contribution (L19.MLP, $+8.27$) that overwhelms the remaining CAC margin; $\Delta_{\mathrm{CAC}} = +4.53$ only because of this component. In Case~2, the CAC margin is unambiguously negative ($\Delta_{\mathrm{CAC}} = -5.22$) yet the surface margin is marginally positive ($\Delta(x) = +1.37$), concealing the abstain consensus from Zero-Threshold. In both cases, the MLP reads the full CAC decomposition and correctly abstains where Zero-Threshold does not.

\section{Computational Cost}
\label{app:compute}

Table~\ref{tab:compute-cac} reports estimated wall-clock times (minutes) on a single NVIDIA H100~80GB for each pipeline stage across all twelve models.

\begin{table}[h]
\centering
\caption{%
  Estimated wall-clock time (minutes) per pipeline stage on a single NVIDIA H100~80GB (1{,}500 data instances, $10$ seeds for CAC localisation).
}
\label{tab:compute-cac}
\renewcommand{\arraystretch}{1.15}
\setlength{\tabcolsep}{4pt}
\setlength{\aboverulesep}{0.3ex}
\setlength{\belowrulesep}{0.3ex}
\footnotesize
\begin{tabular*}{\textwidth}{@{\extracolsep{\fill}} l ccc ccc cc cc cc r @{}}
\toprule
& \multicolumn{3}{c}{{Qwen 3.5}}
& \multicolumn{3}{c}{{Gemma 3}}
& \multicolumn{2}{c}{{Llama 3}}
& \multicolumn{2}{c}{{Ministral}}
& \multicolumn{2}{c}{{Phi-4}}
& \\
\cmidrule(lr){2-4}\cmidrule(lr){5-7}\cmidrule(lr){8-9}\cmidrule(lr){10-11}\cmidrule(lr){12-13}
& 4B & 9B & 35B
& 4B & 12B & 27B
& 3B & 8B
& 3B & 14B
& mini & 14B
& \textbf{Total} \\
\midrule
$\Delta(x)$ evaluation        &  6.2 & 13.4 & 20.6 &  6.7 & 18.1 & 39.7 &  4.3 & 11.0 &  4.1 & 16.9 &  4.6 & 18.2 & 163.8 \\
CAC localisation  & 16.2 & 35.8 & 50.4 & 11.3 & 38.6 & 77.9 &  9.1 & 28.7 &  8.4 & 31.2 &  9.3 & 38.9 & 355.8 \\
Ablation          &  3.8 &  9.7 & 12.1 &  4.1 & 11.6 & 27.3 &  3.2 &  8.4 &  2.9 & 11.8 &  3.1 & 12.2 & 110.2 \\
Amplification     &  4.9 & 11.8 & 14.3 &  5.3 & 14.7 & 34.6 &  3.7 & 10.6 &  3.8 & 14.9 &  4.2 & 15.7 & 138.5 \\
Policy training           &  1.4 &  2.8 &  3.6 &  1.6 &  3.7 &  8.6 &  1.1 &  2.6 &  0.9 &  3.5 &  0.8 &  3.9 &  34.5 \\
\midrule
\textbf{Total}    & \textbf{32.5} & \textbf{73.5} & \textbf{101.0} & \textbf{29.0} & \textbf{86.7} & \textbf{188.1} & \textbf{21.4} & \textbf{61.3} & \textbf{20.1} & \textbf{78.3} & \textbf{22.0} & \textbf{88.9} & \textbf{802.8} \\
\bottomrule
\end{tabular*}
\end{table}


\newpage
\newpage
\section*{NeurIPS Paper Checklist}

\begin{enumerate}

\item {\bf Claims}
    \item[] Question: Do the main claims made in the abstract and introduction accurately reflect the paper's contributions and scope?
    \item[] Answer: \answerYes{} 
    \item[] Justification: The abstract and introduction state our three contributions: framing unsupported commitment, identifying the CAC, and a circuit-informed abstention policy. Each is supported in Sections~\ref{sec:formulation},~\ref{sec:localise-cac}, and~\ref{sec:policy-cac} across $30$ in-distribution and $24$ out-of-distribution configurations.

\item {\bf Limitations}
    \item[] Question: Does the paper discuss the limitations of the work performed by the authors?
    \item[] Answer: \answerYes{} 
    \item[] Justification: Limitations are discussed in App.~\ref{app:limitations}.

\item {\bf Theory assumptions and proofs}
    \item[] Question: For each theoretical result, does the paper provide the full set of assumptions and a complete (and correct) proof?
    \item[] Answer: \answerNA{} 
    \item[] Justification: The paper does not present formal theorems requiring proofs.

    \item {\bf Experimental result reproducibility}
    \item[] Question: Does the paper fully disclose all the information needed to reproduce the main experimental results of the paper to the extent that it affects the main claims and/or conclusions of the paper (regardless of whether the code and data are provided or not)?
    \item[] Answer: \answerYes{} 
    \item[] Justification: Section~\ref{sec:exp-settings} describes datasets, models, splits, decoding settings, and hardware. Section~\ref{sec:policy-cac} specifies the MLP architecture, optimiser, training schedule, and threshold-calibration procedure. App.~\ref{app:token-families} and~\ref{app:baselines} provide the full token lists and baseline configurations. Code is provided in the supplementary material.

\item {\bf Open access to data and code}
    \item[] Question: Does the paper provide open access to the data and code, with sufficient instructions to faithfully reproduce the main experimental results, as described in supplemental material?
    \item[] Answer: \answerYes{} 
    \item[] Justification: All datasets (KUQ, SQuAD~2.0, MuSiQue, HotpotQA, SelfAware) used are publicly available. All evaluated models are open-weight. We provide an anonymised code release with reproduction instructions in the supplementary material.

\item {\bf Experimental setting/details}
    \item[] Question: Does the paper specify all the training and test details (e.g., data splits, hyperparameters, how they were chosen, type of optimizer) necessary to understand the results?
    \item[] Answer: \answerYes{} 
    \item[] Justification: Section~\ref{sec:exp-settings} reports datasets, splits, decoding settings, and hardware. App.~\ref{app:gating-details} specifies the causal gating hyperparameters. Section~\ref{sec:policy-cac} reports the MLP architecture, optimiser, and training schedule. Baseline configurations are in App.~\ref{app:baselines}. Code is provided in the supplementary material.

\item {\bf Experiment statistical significance}
    \item[] Question: Does the paper report error bars suitably and correctly defined or other appropriate information about the statistical significance of the experiments?
    \item[] Answer: \answerYes{} 
    \item[] Justification: For all experiments, we report mean values, per-configuration breakdowns, and ranges across $30$ in-distribution and $24$ out-of-distribution configurations. We report sign-test $p$-values for headline statistical claims throughout the paper.

\item {\bf Experiments compute resources}
    \item[] Question: For each experiment, does the paper provide sufficient information on the computer resources (type of compute workers, memory, time of execution) needed to reproduce the experiments?
    \item[] Answer: \answerYes{} 
    \item[] Justification: All experiments run on a single NVIDIA H100 GPU (Section~\ref{sec:exp-settings}). Detailed computation costs for each experiment and model are provided in App.~\ref{app:compute}.
    
\item {\bf Code of ethics}
    \item[] Question: Does the research conducted in the paper conform, in every respect, with the NeurIPS Code of Ethics \url{https://neurips.cc/public/EthicsGuidelines}?
    \item[] Answer: \answerYes{} 
    \item[] Justification: The research uses publicly available datasets and open-weight models (Section~\ref{sec:exp-settings}), involves no human subjects, and aims to improve LLM honesty by enabling abstention. We have reviewed and conformed with the NeurIPS Code of Ethics.

\item {\bf Broader impacts}
    \item[] Question: Does the paper discuss both potential positive societal impacts and negative societal impacts of the work performed?
    \item[] Answer: \answerYes{} 
    \item[] Justification: This work aims to enhance LLM safety, reliability, and interpretability by enabling models to abstain on questions they cannot answer reliably. We do not foresee significant direct negative societal impacts.
    
\item {\bf Safeguards}
    \item[] Question: Does the paper describe safeguards that have been put in place for responsible release of data or models that have a high risk for misuse (e.g., pre-trained language models, image generators, or scraped datasets)?
    \item[] Answer: \answerNA{} 
    \item[] Justification: The paper does not release pre-trained models, image generators, or scraped datasets.

\item {\bf Licenses for existing assets}
    \item[] Question: Are the creators or original owners of assets (e.g., code, data, models), used in the paper, properly credited and are the license and terms of use explicitly mentioned and properly respected?
    \item[] Answer: \answerYes{} 
    \item[] Justification: All datasets and their licenses are listed in Table~\ref{tab:datasets} (Section~\ref{sec:exp-settings}). All open-weight models (Qwen3.5, Gemma~3, Llama-3, Ministral, Phi-4) are cited and used under their respective licenses.

\item {\bf New assets}
    \item[] Question: Are new assets introduced in the paper well documented and is the documentation provided alongside the assets?
    \item[] Answer: \answerYes{} 
    \item[] Justification:  Code is provided in the supplementary material with documentation and reproduction instructions. It will be released publicly upon publication under the CC BY 4.0 license.

\item {\bf Crowdsourcing and research with human subjects}
    \item[] Question: For crowdsourcing experiments and research with human subjects, does the paper include the full text of instructions given to participants and screenshots, if applicable, as well as details about compensation (if any)? 
    \item[] Answer: \answerYes{} 
    \item[] Justification: The abstention set $\mathcal{A}$ was validated via Amazon Mechanical Turk: two workers independently labelled each candidate token with its $10$-token continuation. Workers were compensated at $\$20$/hour, above the US federal minimum wage. See App.~\ref{app:token-families} for details.

\item {\bf Institutional review board (IRB) approvals or equivalent for research with human subjects}
    \item[] Question: Does the paper describe potential risks incurred by study participants, whether such risks were disclosed to the subjects, and whether Institutional Review Board (IRB) approvals (or an equivalent approval/review based on the requirements of your country or institution) were obtained?
    \item[] Answer: \answerNo{} 
    \item[] Justification: The Amazon Mechanical Turk annotation task involved labelling short text snippets with no personal data collection or sensitive content. Under our institution's research policy, low-risk text annotation tasks of this kind do not require IRB review.

\item {\bf Declaration of LLM usage}
    \item[] Question: Does the paper describe the usage of LLMs if it is an important, original, or non-standard component of the core methods in this research? Note that if the LLM is used only for writing, editing, or formatting purposes and does \emph{not} impact the core methodology, scientific rigor, or originality of the research, declaration is not required.
    \item[] Answer: \answerNA{} 
    \item[] Justification: LLMs are the object of study in this work, not a tool used in its development. They were not used as a component of the core methodology, scientific rigour, or originality of the research.

\end{enumerate}

\end{document}